\documentclass{article}
\usepackage{amsmath,bbm}
\usepackage{algorithm}
\usepackage{algorithmic}
\usepackage{multirow,graphicx}
\usepackage{rotating}
\usepackage{color,colortbl}
\usepackage{enumitem}
\usepackage{tikz}
\usetikzlibrary{arrows}

\usetikzlibrary{external}
\tikzset{external/system call={pdflatex \tikzexternalcheckshellescape -halt-on-error 
-interaction=batchmode -jobname "\image" "\texsource" &&
pdftops -eps "\image".eps}}
\definecolor{gray}{gray}{0.9}

\newcommand{\argmax}{\operatornamewithlimits{argmax}}

\title{An Extensive Evaluation of Filtering Misclassified Instances in Supervised Classification Tasks}
\author{Michael R. Smith and Tony Martinez\\
\small Department of Computer Science, Brigham Young University, Provo, UT 84602 USA\\
\small \texttt{msmith@axon.cs.byu.edu, martinez@cs.byu.edu}}

\begin{document}

% \begin{frontmatter}

% \title{An Extensive Evaluation of Filtering Misclassified Instances in Supervised Classification Tasks}
% \author{Michael R. Smith}
% \ead{msmith@axon.cs.byu.edu}
% \author{Tony Martinez}
% \ead{martinez@cs.byu.edu}
% \address{Department of Computer Science, Brigham Young University, Provo, UT 84602 USA}
\date{}
\maketitle

\begin{abstract}
% As many real-world data sets are noisy, not all instances in a data set are equally beneficial for inferring a model of the data set and some can even be detrimental.
% For example, real-world data sets are often noisy.
% Many methods exist that identify outliers and noisy instances.
Removing or filtering outliers and mislabeled instances prior to training a learning algorithm has been shown to increase classification accuracy.
A popular approach for handling outliers and mislabeled instances is to remove any instance that is misclassified by a learning algorithm.
However, an examination of which learning algorithms to use for filtering as well as their effects on multiple learning algorithms over a large set of data sets has not been done.
Previous work has generally been limited due to the large computational requirements to run such an experiment, and, thus, the examination has generally been limited to learning algorithms that are computationally inexpensive and using a small number of data sets.
In this paper, we examine 9 learning algorithms as filtering algorithms as well as examining the effects of filtering in the 9 chosen learning algorithms on a set of 54 data sets.
In addition to using each learning algorithm individually as a filter, we also use the set of learning algorithms as an ensemble filter and use an adaptive algorithm that selects a subset of the learning algorithms for filtering for a specific task and learning algorithm.
We find that for most cases, using an ensemble of learning algorithms for filtering produces the greatest increase in classification accuracy.
We also compare filtering with a majority voting ensemble.
The voting ensemble significantly outperforms filtering unless there are high amounts of noise present in the data set.
Additionally, we find that a majority voting ensemble is robust to noise as filtering with a voting ensemble does \textit{not} increase the classification accuracy of the voting ensemble.
% % A common approach for filtering instances is to remove instances that are misclassified. 
% In this paper, detrimental instances in a data set are identified using a set of learning algorithms (a \textit{filter set}) trained on the data set.
% We investigate using filter sets to filter training data and present an adaptive algorithm that selects a filter set for a specific task and learning algorithm.
% % We compare filter sets with other filtering approaches on a set of 52 data sets and 9 learning algorithms.
% We find that adaptively creating a filter set for a particular task and learning algorithm significantly increases classification accuracy over other filtering methods.
% On a set of 52 data sets using 6 learning algorithms, the adaptive approach for selecting the filter set increases the average classification accuracy from 82\% to 87\% and significantly outperforms an ensemble.
\end{abstract}

% \begin{keyword}
%% keywords here, in the form: keyword \sep keyword

%% MSC codes here, in the form: \MSC code \sep code
%% or \MSC[2008] code \sep code (2000 is the default)

% \end{frontmatter}

\section{Introduction}
The goal of supervised machine learning is to infer an accurate generalizing function $\mathcal{F}: X \mapsto Y$ from a set of input feature vectors $X =\{x_1, x_2, \dots ,x_n\}$ and a corresponding set of of label vectors $Y =\{y_1, y_2, \dots, y_n\}$.
The quality of the inferred function $\mathcal{F}$ by a learning algorithm is dependent on the quality of the data used for training.
Many real-world data sets are often noisy where the noise in a data set can be label noise and/or attribute noise \cite{Zhu2004}.
The focus of this paper is on label noise.
Noise arises from various sources such as subjectivity, human errors, and sensor malfunctions.
As such, it is important to take the possibility of label noise into account when inferring a model of the data.
Much previous work has examined the effects of class noise and how to handle it.
As many real-wold data sets are inherently noisy, most learning algorithms are designed to tolerate a certain degree of noise by avoiding overfitting the training data.
There are two general approaches for handling class noise: 1) creating learning algorithms that are robust to noise such as the C4.5 algorithm for decision trees \cite{Quinlan1993} and 2) preprocessing the data prior to inferring a model of the data such as filtering \cite{Wilson1972,Brodley1999} or correcting \cite{Teng2003} noisy instances.
In this work, we specifically examine handling noise by filtering.

Previous work has generally examined filtering in a limited context using a single or very few learning algorithms and/or using a limited number of data sets.
This may be in part due to the extra computational requirement to first filter a data set and then infer a model of the data using the filtered data set.
As such, previous work has generally limited itself to investigating relatively fast learning algorithms such as decision trees \cite{John95} and nearest-neighbor algorithms \cite{Tomek1976,Wilson2000}.
In addition, filtering for instance-based learning algorithms was motivated in part to reduce the number of instances that have to be stored and because instance-based learning algroithms are more sensitive to noise than other learning algorithms.
Also, most previous work artificially added noise to the data set to show that filtering, weighting, or cleaning the data set is beneficial.
In this work, we examine filtering misclassified instances over a set of 54 data sets and 9 learning algorithms \textit{without} adding artificial noise.
The artificial noise was added in previous work to show that filtering/weighting/cleaning provided significant improvements with noisy data sets.
Within the context of the benefits of filtering established by the previous work, we show how filtering affects data sets without adding artificial noise to a data set.
This also avoids making assumptions about the generation of the noise which may or may not be accurate.
We also compare filtering with a voting ensemble with a diverse set of base classifiers.

The insights provided shed light on which learning algorithms are beneficial for filtering and which learning algorithms are the most robust to noise.
Using a larger number of data sets allows for more statistical confidence in the results than if only a small number of data sets are used.
We find that using an ensemble filter achieves significantly higher classification accuracy than using a single learning algorithm filter.
We also find that, in general, a voting ensemble is robust to noise and achieves significantly higher classification accuracy trained on unfiltered data than a single learning algorithm trained on filtered data.
On data sets with higher percentages of inherent noisy instances, however, using an ensemble filter achieves higher classification accuracy than a voting ensemble for some learning algorithms.
And surprisingly, training a voting ensemble on filtered training data significantly \textit{decreases} classification accuracy compared to training a voting ensemble on unfiltered training data.

In the next section, we present previous work for handling noise in supervised classification problems.
A mathematical motivation for filtering misclassified instances is presented in Section \ref{section:math}.
We then present our experimental methodology in Section \ref{section:methodology} followed by a presentation of the results in Section \ref{section:results}.
In Section \ref{section:conclusions} we provide conclusions and directions for future work.

\section{Related Work}
\label{section:relatedWork}
As many real-wold data sets are inherently noisy, most learning algorithms are designed to tolerate a certain degree of noise.
Typically, learning algorithms are designed to be somewhat robust to noise by making a trade-off between the complexity of the inferred model and optimizing the inferred function on the training data to prevent overfit.
Some techniques to avoid overfit include early stopping using a validation set, pruning (such as in the C4.5 algorithm for decision trees \cite{Quinlan1993}), or regularization by adding a complexity penalty to the loss function \cite{bishop2006pattern}.
Further, some learning algorithms have been adapted specifically to better handle label noise.
For example, noisy instances are problematic for boosting algorithms \cite{Schapire1990, Freund1995} where more weight is placed upon misclassified instances, which often include mislabeled and noisy instances.
To address this, Servedio \cite{Servedio2003} presented a boosting algorithm that does not place too much weight on any single training instance. 
For support vector machines, Collobert et al. \cite{Collobert2006} use the ramp-loss function to place a bound on the maximum penalty for an instance that lies on the wrong side of the margin.
Lawrence and Sch\"{o}lkopf \cite{Lawrence2001} explicitly model the possibility that an instance is mislabeled using a generative model and then use expectation maximization to update the probability that an instance is mislabeled.

Preprocessing the data set is another approach that explicitly handles label noise.
This can be done by removing noisy instances, weighting the instances, or correcting incorrect labels.
All three approaches first attempt to identify which instances are noisy by various criteria.
Filtering noisy instances has received much attention and has generally resulted in an increase in classification accuracy \cite{Gamberger2000,Smith2011}.
One frequently used filtering technique removes any instance that is misclassified by a learning algorithm \cite{Wilson1972} or set of learning algorithms \cite{Brodley1999}.
Verbaeten and Van Assche \cite{Verbaeten2003} further pursued the idea of using an ensemble for filtering using ideas from boosting and bagging.
Other approaches use learning algorithm heuristics to remove noisy instances.
Segata et al. \cite{Segata2009} remove instances that are too close or on the wrong side of the decision surface generated by a support vector machine.
Zeng and Martinez \cite{Xinchuan2003} remove instances while training a neural network that have a low probability of being labeled correctly where the probability is calculated using the output from the neural network.
Filtering has the potential downside of discarding useful instances.
However, it is assumed that there are significantly more non-noisy instances and that throwing away a few correct instances with the noisy instances will not have a negative impact on a large data set.

Weighting the instances in a training set has the benefit of not discarding any instances.
Rebbapragada and Brodley \cite{Rebbapragada2007} weight the instances using expectation maximization to cluster instances that belong to a pair of the classes.
The probabilities between classes for each instances is compiled and used to weight the influence of each instance.
Smith and Martinez \cite{Smith_RIDL} examine weighting the instances based on their probability of being misclassified.

Similar to weighting the training instances, data cleaning does not discard any instances, but rather strives to correct the noise in the instances.
As in filtering, the output from a learning algorithm has been used to clean the data.
Automatic data enhancement \cite{zeng.ida2001} uses the output from a neural network to correct the label for training instances that have a low probability of being correctly labeled.
Polishing \cite{Teng2000,Teng2003} trains a learning algorithm (in this case a decision tree) to predict the value for each attribute (including the class).
The predicted (i.e.~correct) attribute values for the instances that increase generalization accuracy on a validation set are used instead of the uncleaned attribute values.

We differ from the related work in that we do not add artificial noise to the data sets when we examine filtering.
Thus, we avoid making any assumptions about the noise source and focus on the noise inherent in the data sets.
We also examine the effects of filtering on a larger set of learning algorithms and data sets providing more significance to the generality of the results.

% \section{Motivation for Removing Misclassified Instances}
\section{Modeling Class Noise in a Discriminative Model}
\label{section:math}
\begin{figure*}
\begin{center}
\includegraphics[scale=0.35]{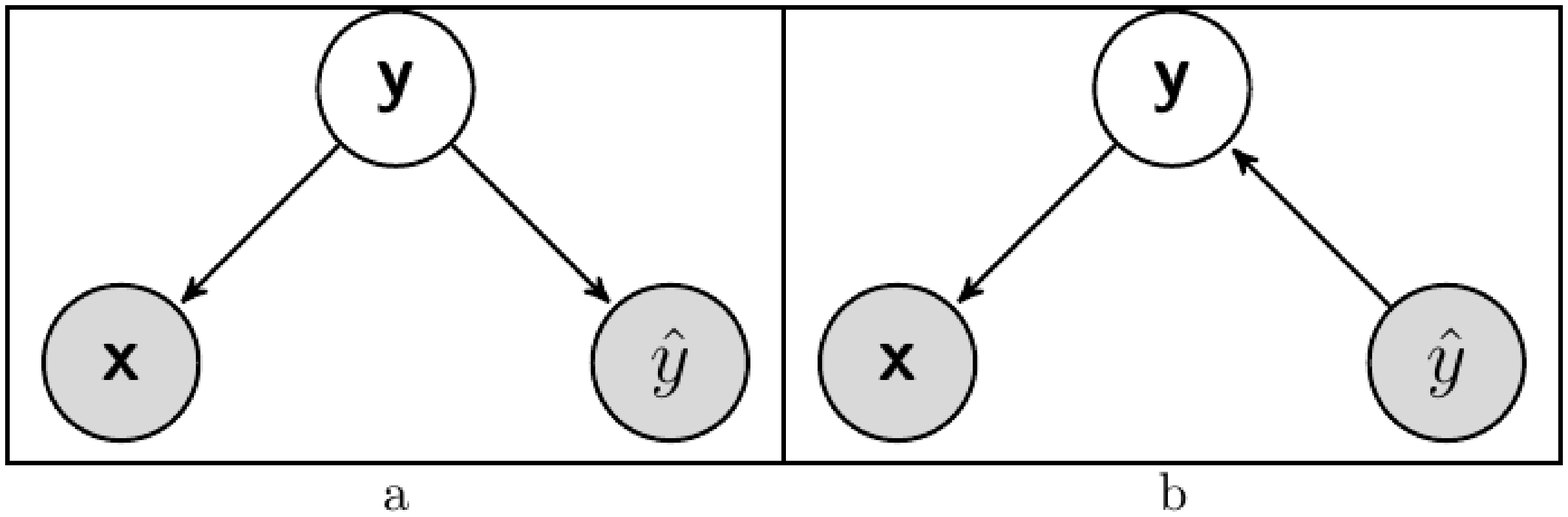}
\caption{Graphical model of the generative probabilistic model proposed by Lawrence and Sch\"{o}lkopf \cite{Lawrence2001}.}
\label{figure:generativeModel}
\end{center}
\end{figure*}

Lawrence and Sch\"{o}lkopf \cite{Lawrence2001} proposed to model a data set probabilistically using a generative model that models the noise process.
They assume that the joint distribution $p(x,y,\hat{y})$ (where $x$ is the set of input features, $\hat{y}$ is the possibly noisy class label given in the trianing set, and $y$ is the actual unkown class label) is factorized as $p(\hat{y}|y)p(x|y)p(y)$ as shown in Figure \ref{figure:generativeModel}a.
However, since modeling the prior distribution of the unobserved random variable $y$ is not feasible, it is more practical to estimate the prior distribution of $p(\hat{y})$ with some assumptions about the class noise as shown in Figure \ref{figure:generativeModel}b.

Here, we follow the premise of Lawrence and Sch\"{o}lkopf by explicitly modeling the possibility that an instance is misclassified.
Rather than using a generative model, though, we use a discriminative model since we are focusing on classification tasks and do not require the full joint distribution.
Also, discriminative models have been shown to yield better performance on classification tasks \cite{Ng+Jordan:2001}.

\begin{figure*}
\begin{center}
\includegraphics[scale=0.35]{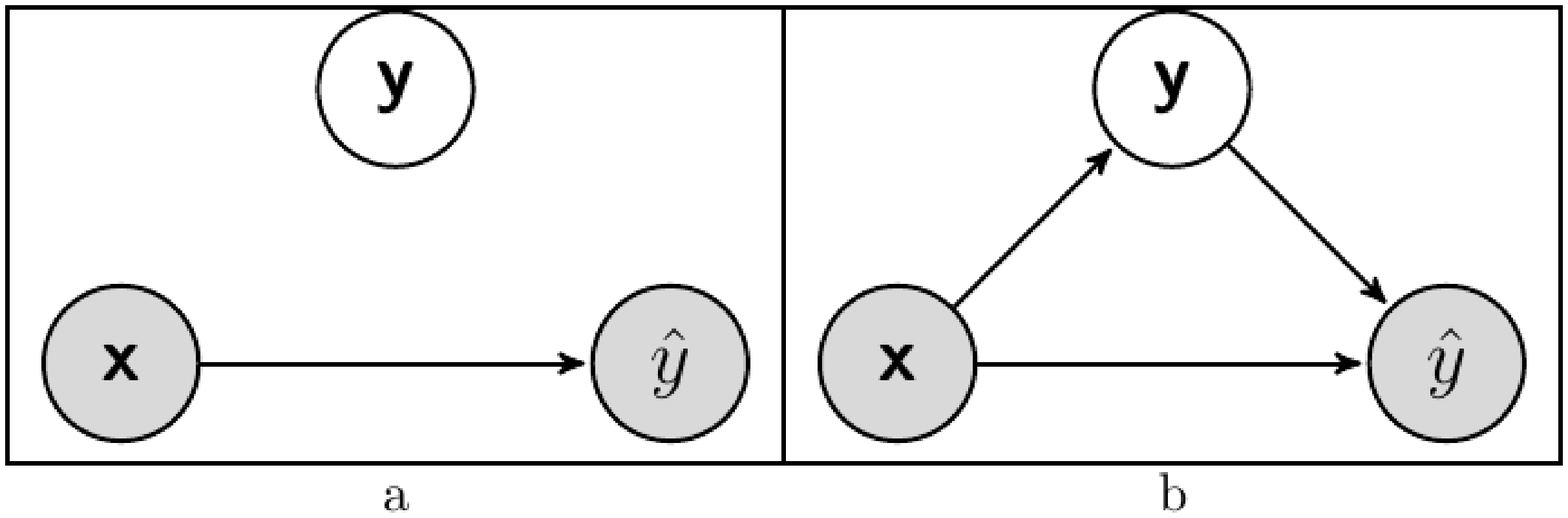}
% \begin{tabular}{|c|c|}
% \hline
% \begin{tikzpicture}[->,>=stealth',shorten >=1pt,auto,node distance=2.5cm,
%   thick,main node/.style={circle,fill=darkgray!20,draw,font=\sffamily\Large\bfseries,minimum size=10mm},plain node/.style={circle,draw,font=\sffamily\Large\bfseries,minimum size=10mm}]
%   \node[plain node] (7) {y};
%   \node[main node] (8) [below left of=7] {x};
%   \node[main node] (9) [below right of=7] {$\hat{y}$};
% 
%   \path[every node/.style={font=\sffamily\small}]
%     (8) edge node [right] {} (9);
% \end{tikzpicture}
% &
% \begin{tikzpicture}[->,>=stealth',shorten >=1pt,auto,node distance=2.5cm,
%   thick,main node/.style={circle,fill=darkgray!20,draw,font=\sffamily\Large\bfseries,minimum size=10mm},plain node/.style={circle,draw,font=\sffamily\Large\bfseries,minimum size=10mm}]
% 
%   \node[plain node] (10) {y};
%   \node[main node] (11) [below left of=10] {x};
%   \node[main node] (12) [below right of=10] {$\hat{y}$};
% 
%   \path[every node/.style={font=\sffamily\small}]
%     (11) edge node [left] {} (10)
%         edge node [right] {} (12)
%     (10) edge node [right] {} (12);
% \end{tikzpicture}
% 
% \\\hline
% \multicolumn{1}{c}{a} & \multicolumn{1}{c}{b}\\
% \end{tabular}
\caption{Graphical representation of a discriminative probabilistic model for a) $p(\hat{y}|x)p(x)$ and b) $p(\hat{y}|x,y)p(y|x)p(x)$.}
\label{figure:discriminativeModel}
\end{center}
\end{figure*}

Let $T$ be a training set composed of instances $\langle x_i, \hat{y}_i\rangle$ drawn i.i.d. from the underlying data distribution $\mathcal{D}$.
Each instance is composed of an input vector $x_i$ with a corresponding possibly noisy label vector $\hat{y}_i$.
Given the training data $T$, a learning algorithm generally seeks to find the most probable hypothesis $h$ that maps each $x_i \mapsto \hat{y}_i$.
For supervised classification problems, most learning algorithms maximize $p(\hat{y}_i|x_i,h)$ for all instances in $T$.
This is shown graphically in Figure \ref{figure:discriminativeModel}a where the probabilities are estimated using a discriminative approach such as a neural network or a decision tree to infer a hypothesis of the data.
Using Bayes' rule and decomposing $T$ into its individual constituent instances, the maximum a posteriori hypothesis is:
\begin{align}
\argmax_{h \in \mathcal{H}} p(h|T)&=\frac{p(T|h)p(h)}{p(T)} \nonumber\\
&=\prod_i p(x_i,\hat{y}_i|h)p(h)\nonumber\\
\argmax_{h \in \mathcal{H}} p(h|T)&=\prod_i p(\hat{y}_i|x_i,h) p(x_i|h)p(h). \label{eq:1}
\end{align}
In Equation \ref{eq:1}, the MAP hypothesis $h$ is found by finding a global optima where all instances are included in the optimization problem.
However, noisy instances are often detrimental for finding the global optima since they are not representative of the true (and unknown) underlying data distribution $\mathcal{D}$.
The possibility of label noise is not explicitly modeled in this form--completely ignoring $y_i$.
Thus, label noise is generally handled by avoiding overfit such that more probable, simpler hypotheses are preferred ($p(h)$).
The possibility of label noise can be modeled explicitly by including the latent random variable $y_i$ with $x_i$ and $\hat{y}_i$.
Thus, an instance is the triplet $\langle x_i, \hat{y}_i, y_i\rangle$ and a supervised learning algorithm seeks to maximize $p(\hat{y}_i|x_i,y,h)$--modeled graphically in Figure \ref{figure:discriminativeModel}b.
Using the model in Figure \ref{figure:discriminativeModel}b, the MAP hypothesis becomes:
\begin{align}
\argmax_{h \in \mathcal{H}} p(h|T)&=\prod_i p(x_i,y_i,\hat{y}_i|h)p(h)\nonumber\\
&=\prod_i p(\hat{y}_i|x_i,y_i,h)p(y_i|x_i,h) p(x_i|h)p(h).\label{eq:noise}
\end{align}
Equation \ref{eq:noise} shows that for an instance $x_i$, the probability of an observed class label ($p(\hat{y}_i|x_i,y_i,h)$) should be weighted by the probability of the actual class ($p(y_i|x_i,h)$).
We now show a method to estimate $p(y|x,h)$.

For filtering as a preprocessing step, we want to calculate $p(y_i|\hat{y_i},x_i)$ and remove instances that have a low probability of $y_i$ being the same as $\hat{y_i}$.
Using a discriminative model $h$ trained on $T$, we can calculate $p(y_i|\hat{y}_i,x_i,h)$ as
\begin{align}
p(y_i|\hat{y_i},x_i,h)&= p(y_i|\hat{y}_i, h)p(\hat{y}_i|x_i,h). \nonumber
% &\approx p(\hat{y}_i|x_i,h). \nonumber
\end{align}
Since the quantity $p(y|\hat{y}_i)$ is unknown, $p(y_i|\hat{y}_i,x_i,h)$ can be approximated as $p(\hat{y}_i|x_i,h)$ assuming that $p(y_i|\hat{y}_i)$ is represented in $h$.
In other words, the inferred discriminative model is able to model if one class label is more likely than another class label given an observed noisy label.
Otherwise, all class labels are assumed to be equally likely given an observed label.
Thus, $p(y_i|\hat{y}_i,x_i,h)$ can be approximated by finding the class distributions for a given $x_i$ from an inferred discriminative model.
That is, after training a learning algorithm on $T$, the class distribution for an instance $x_i$ can be calculated based on the output from the learning algorithm.
As shown in Equation \ref{eq:1}, $p(\hat{y}_i|x_i,h)$ is found naturally through a derivation of Bayes' law.
The quantity $p(\hat{y}_i|x_i,h)$ is the maximum likelihood of an instance given a hypothesis $h$ which a learning algorithm tries to maximize for each instance.
Further, the dependence on $h$ can be removed by summing over all possible hypotheses $h$ in $\mathcal{H}$ and multiplying each $p(\hat{y}_i|x_i,h)$ by $p(h)$:
\begin{equation}
p(y_i|\hat{y}_i,x_i)\approx p(\hat{y}_i|x_i) = \sum_{h\in\mathcal{H}} p(\hat{y}_i|x_i,h)p(h). \label{eq:sumOverH}
\end{equation}
This formulation is infeasible though because 1) it is not practical (or possible) to sum over the set of all hypotheses, 2) calculating $p(h)$ is non-trivial, and 3) not all learning algorithms produce a probability distribution.
These limitation make probabilistic generative models attractive, such as the kernel Fisher discriminant algorithm \cite{Lawrence2001}.
However, for classification tasks, generative models generally have a higher asymptotic error than discriminative models \cite{Ng+Jordan:2001}.

% Also, to lessen the dependence on a single hypothesis, multiple hypotheses are used and $p(h)$ is non-zero for the chosen hypotheses and zero for all other hypotheses.
% If only a single learning algorithm is used then only the misclassified instances are removed.

\section{Methodology}
\label{section:methodology}

\begin{figure*}
\begin{center}
\input{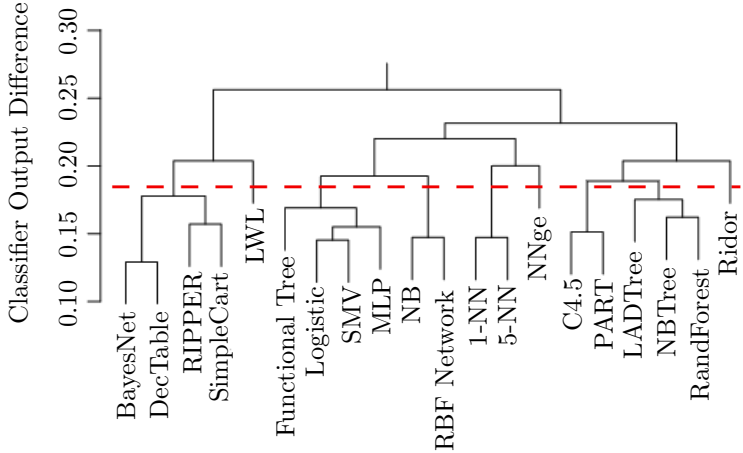}
\caption{Dendrogram of the considered learning algorithms clustered using unsupervised metalearning based on their classifier output difference.}
\label{figure:COD}
\end{center}
\end{figure*}
The first step for filtering is to determine $p(y_i|\hat{y}_i,x_i,h)$ for each instance.
Given that a number of different techniques could be employed to estimate $p(y_i|\hat{y}_i,x_i,h)$, we conduct an extensive evaluation of filtering misclassified instances using a diverse set of learning algorithms.
The diversity of the learning algorithm refers to the learning algorithms not having the same classification for all of the instances and is determined using unsupervised meta-learning (UML) \cite{Lee2011}.
UML first uses Classifier Output Difference (COD) \cite{Peterson2005} to measure the diversity between learning algorithms.
COD measures the distance between two learning algorithms as the probability that the learning algorithms make different predictions.
UML then clusters the learning algorithms based on their COD scores with hierarchical agglomerative clustering.
We considered 20 learning algorithms from Weka with their default parameters \cite{weka2009}.
The resulting dendrogram is shown in Figure \ref{figure:COD}, where the height of the line connecting two clusters corresponds to the distance (COD value) between them.
A cut-point of 0.18 was chosen to create 9 clusters and a representative algorithm from each cluster was used to create a diverse set of learning algorithms.
The learning algorithms that were used are listed in Table \ref{table:LA}.

\begin{table}
\caption{Set of learning algorithms used for filtering.}
\label{table:LA}
\begin{center}
\setlength{\tabcolsep}{3pt}
\begin{tabular}{ll}
\multicolumn{2}{c}{Learning Algorithms}\\
\hline
*& Multilayer Perceptron trained with Back Propagation (MLP) \\
*& Decision Tree (C4.5) \cite{Quinlan1993} \\
*& Locally Weighted Learning (LWL) \\
*& 5-Nearest Neighbors (5-NN) \\
*& Nearest Neighbor with generalization (NNge) \\
*& Na\"{i}ve Bayes (NB) \\
*& RIpple DOwn Rule learner (RIDOR) \\
*& Random Forest (RandForest) \\
*& Repeated Incremental Pruning to Produce Error Reduction (RIPPER) \\
\end{tabular}
\end{center}
\end{table}

We investigate filtering using the learning algorithms shown in Table \ref{table:LA}.
Since not all learning algorithms produce a probability distribution, the indicator function $\mathbbm{1}(h(x_i)=\hat{y}_i)$ is used in this paper instead of $p(\hat{y}_i|x_i,h)p(h)$, thus, removing misclassified instances.
Each learning algorithm first filters misclassified instances and then infers a model of the data using the filtered data set.
We also examine using an ensemble filter--removing instances that are misclassified by different percentages of the 9 learning algorithms.
The ensemble filter more closely approximates $p(y|\hat{y}_i,x_i)$ from Equation \ref{eq:sumOverH} since it sums over a set of learning algorithms (which in this case were chosen to be diverse and represent a larger subset of the hypothesis space $\mathcal{H}$) lessening the dependence on a single hypothesis $h$.
For the ensemble filter, $p(y|\hat{y}_i,x_i)$ is estimated using a subset of learning algorithms $\mathcal{L}$:
\begin{align}
p(y|\hat{y}_i,x_i) \approx p(\hat{y}_i|x_i,\mathcal{L}) 
&\approx \frac{1}{|\mathcal{L}|} \sum_{j=1}^{|\mathcal{L}|} p(\hat{y}_i|x_i, l_j(T))
\end{align}
where $l_j(T)$ is the hypothesis from the $j^{th}$ learning algorithm trained on training set $T$.
From Equation \ref{eq:sumOverH}, $p(h)$ is estimated as $\frac{1}{|\mathcal{L}|}$ for the $j^th$ hypothesis generated from training the learning algorithms in $\mathcal{L}$ on $T$ and as zero for all of the other hypotheses in $\mathcal{H}$.
Also, $p(\hat{y}_i|x_i, l_j(T))$ is estimated using the indicator function since not all learning algorithms produce a probability distribution over the output classes.
Set up as such, the ensemble filter counts how many times an instance is misclassified by a set of learning algorithms.
Brodley and Friedl \cite{Brodley1999} examined an ensemble of three learning algorithms on five data sets with artificially generated noise inserted into the data sets.
In this paper, we examine an ensemble filter, removing instances that are misclassified by 50, 70, and 90 percent of the learning algorithms in the ensemble.
One of the problems of using an ensemble filter is having to choose the percentage of learning algorithms that misclassify an instance for filtering.
For the results, we report the accuracy from the percentage that produces the highest accuracy using 5 by 10-fold cross-validation to choose the best percentage for each data set.
This method highlights the impact of using an ensemble filter, however, in practice a validation set is often used to determine the percentage that would be used.

In addition, we also examine using an adaptive filtering approach that iteratively adds a learning algorithm to a set of filtering learning algorithms by selecting the learning algorithm from a set of candidate learning algorithms $\mathcal{L}$ that produces the highest classification accuracy on a validation set when added to the set of learning algorithms used for filtering, as shown in Algorithm \ref{figure:greedySearch}.
The function $runLA(F)$ trains a learning algorithm on a data set using the filter set $F$ to filter the instances and returns the accuracy of the learning algorithm on a validation set.
As with the ensemble filter, instances are removed that are misclassified by a given percentage of the filtering learning algorithms.
The idea is to choose an optimal subset of learning algorithms through a greedy search of the candidate filtering algorithms.
% Using this subset of algorithms could then lead to insights about which algorithms are the most appropriate for filtering.
For the results, we report the accuracy from the percentage that produces the highest accuracy using 5 by 10-fold cross-validation to choose the best percentage for each data set.

\begin{algorithm}[tb]
\caption{Adaptively constructing a filter set.}
\label{figure:greedySearch}
\begin{algorithmic}[1]
\STATE Let $F$ be the filter set used for filtering and $\mathcal{L}$ be the set of candidate learning algorithms for $F$. % filtering.
\STATE Initialize $F$ to the empty set: $F \gets \{\}$ %\Comment{Filter set}
\STATE Initialize the current accuracy to the accuracy from an empty filter set: $currAcc\gets runLA(\{\})$. $runLA(F)$ returns the accuracy from a learning algorithm trained on a data set filtered with $F$.\label{line:init}%\Comment{Set currAcc to no filtering}
\WHILE{$\mathcal{L} \neq \{\}$} % \&\& ($currAcc > prevAcc$)}
   \STATE $bestAcc \gets currAcc$; $bestLA \gets null$;
   \FORALL{$g \in \mathcal{L}$} %\Comment{Try all remaining algorithms in $\mathcal{G}$}
       \STATE $tempF \gets F + g$; $acc\gets runLA(tempF)$;
       \IF {$acc > bestAcc$}
          \STATE $bestAcc \gets acc$; $bestLA \gets g$;
       \ENDIF
   \ENDFOR
   \IF{$bestAcc > currAcc$}
      \STATE $\mathcal{L} \gets \mathcal{L} - bestLA$; $F \gets F + bestLA$;  $currAcc \gets bestAcc$;
   \ELSE
      \STATE break;
   \ENDIF
\ENDWHILE
 
\end{algorithmic}
\end{algorithm}

Each method for filtering is evaluated using 5 by 10-fold cross-validation (running 10-fold cross validation 5 times, each time with a different seed to partition the data).
We examine filtering using the 9 chosen learning algorithms on a set of 47 data sets from the UCI data repository and 7 non-UCI data sets \cite{Thomson1996,Salojärvi05,promise,gemlrep}.
For filtering, we examine two methods for training the filtering algorithms:
1) removing the misclassified instances when trained on the entire training set and 
2) using cross-validation on the training set that removes instances that are misclassified in the validation set. %  partitioning the training set into a training set for the filtering algorithm and only removing instances that are misclassified from the training test set.
The number of folds for using cross-validation for the training set was set to 2, 3, 4, and 5.
Table \ref{table:dataSets} shows the data sets used in this study organized according to the number of instances, number of attributes, and attribute type.
The non-UCI data sets are in bold.
\begin{table}
\centering
\caption{Datasets used organized by number of instances (\# Ins), number of attributes, and attribute type.}
\begin{tabular}{| c | c | c | c | c |}
\hline
\multirow{2}{*}{\# Ins}& \multirow{2}{*}{\# Attributes} & \multicolumn{3}{| c |}{Attribute Type} \\
\cline{3-5} 
 &  & Categorical & Numerical & Mixed \\
\hline
\multirow{4}{*}{\begin{sideways}$M <$ 100 \end{sideways}} & \multirow{2}{*}{$k <$ 10} & Contact Lenses & & Post-Operative \\
& &  & & \textbf{cm1\_req} \\
\cline{2-5}
& \multirow{2}{*}{$10 < k < 100$} & Lung Cancer & \textbf{desharnais} & Labor \\
& & & & \textbf{Pasture} \\
\hline
\multirow{18}{*}{\begin{sideways}$100<M<1000$ \end{sideways}} & \multirow{6}{*}{$k < 10$} & Breast-w & Iris & Badges 2 \\
& & Breast Cancer & Ecoli & Teaching- \\
& & & Pima Indians & Assistant \\
& & & Glass & \\
& & & Bupa &  \\
& & & Balance Scale &  \\
\cline{2 - 5}
& \multirow{10}{*}{$10 < k < 100$} & Audiology & Ionosphere & Annealing \\
& & Soybean(large) & Wine & Dermatology \\
& & Lymphography & Sonar & Credit-A \\
& & Congressional- & Heart-Statlog & Credit-G \\
& & Voting Records & \textbf{ar1} & Horse Colic \\
& & Vowel & & Heart-c \\
& & Primary-Tumor & & Hepatitis \\
& & Zoo & & Autos \\
& & & & Heart-h \\
& & & & \textbf{eucalyptus} \\
\cline{2-5}
& \multirow{2}{*}{$k > 100$} & & \textbf{AP\_Breast-} & Arrhythmia \\
& & & \textbf{Uterus} & \\
\hline
\multirow{7}{*}{\begin{sideways}$ 1000 < M < 10000$ \end{sideways}} & \multirow{2}{*}{$k < 10$} & Car Evaluation & Yeast &  \\
& & Titanic & & \\
\cline{2-5}
& \multirow{5}{*}{$k < 100$} &  & Waveform-5000 & Thyroid- \\
& &  & Segment & (sick \& \\
& &  & Spambase & hypothyroid) \\
& &  & Ozone level- &  \\
& &  & Detection &  \\
\hline
\multirow{5}{*}{\begin{sideways}$ M > 10000$ \end{sideways}} & \multirow{2}{*}{$k < 10$} & Nursery & MAGIC &  \\
 & & & Telescope & \\
\cline{2-5}
 & \multirow{3}{*}{$k < 100$} & Chess- & & \textbf{Eye-} \\
& & (King-Rook vs. &  & \textbf{movements} \\
& &  King-Pawn) & &  \\
\hline
\end{tabular}
\label{table:dataSets}
\end{table}
Statistical significance between pairs of algorithms is determined using the Wilcoxon signed-ranks test as suggested by Dem\v{s}ar \cite{Demsar2006}.
We emphasize the extensive nature of this evaluation:
\begin{enumerate}
 \item Filtering is examined on 9 diverse learning algorithms.
 \item 9 diverse learning algorithms are examined as misclassification filtering techniques.
 \item In addition to the single algorithm misclassification filters, an ensemble filter and an adaptive filter are examined.
 \item Each filtering method is examined on a set of 54 data sets using 5 by 10-fold cross-validation.
 \item Each filtering method is examined on the entire training set as well as using 2-, 3-, 4-, and 5-fold cross-validation.
\end{enumerate}

\section{Results}
\label{section:results}
In this section, we present the results from filtering the 54 data sets using a biased filter (the same learning algorithm to filter misclassified instances is used to infer a model of the data), an ensemble filter, and the adaptive filter.
Except for the adaptive filter, we find that using cross-validation on the training set for filtering resulted in a lower accuracy (and often significantly lower) accuracy than using the entire training set and, as such, the following results for the biased filter and the ensemble filter are from using the entire training set for filtering rather than using cross-validation.
We first show how filtering affects each learning algorithm in Section \ref{section:filteringResults}.
Next, we examine using a set of data set measures to determine when filtering is the most effective in Section \ref{section:filteringAnalysis}.
Our results suggest that using an ensemble filter in all cases produces the best results.
In Section \ref{section:filteringVsEnsemble}, we then compare filtering with a voting ensemble and show that a voting ensemble is preferable to filtering.

\subsection{Filtering Results}
\label{section:filteringResults}
The filtering results are summarized in Table \ref{table:summary}--showing the average classification accuracy for each learning algorithm and filtering algorithm pair\footnote{The NNge learning algorithm did not finish running two data sets: eye-movements and Magic telescope.
RIPPER did not finish on the lung cancer data set.
In these cases, the data sets are omitted from the presented results.
As such, NNge was evaluated on a set of 52 data sets and RIPPER was evaluated on a set of 53 data sets.}.
The values in bold represent those that are a statistically significant improvement over not filtering.
The results of the statistical significance tests for each of the learning algorithms is provided in Tables \ref{table:sigBackprop}-\ref{table:sigRIPPER} in \ref{section:sigTests}.
The results are summarized below.

\begin{table}
\caption{Summary of filtering using the same learning algorithm to filter misclassified instances and to infer a model of the data, an ensemble filter, and the adaptive filter.
For all learning algorithms, the ensemble filter significantly increases the classification accuracy.}
\setlength{\tabcolsep}{3.4pt}
\begin{tabular}{lccccccccc}
 & MLP & C4.5 & IB5 & LWL & NB & NNge & RF & Rid & RIP \\
\hline
Orig & 81.74 & 80.80 & 79.91 & 72.80 & 76.94 & 80.14 & 82.28 & 79.90 & 79.76 \\
Biased & 81.72 & 80.75 & 79.53 & 70.91 & 75.88 & 80.34 & 82.14 & 79.02 & 79.87 \\
Ensemble & \textbf{83.40} & \textbf{81.61} & \textbf{80.85} & \textbf{73.48} & \textbf{78.92} & \textbf{82.21} & \textbf{82.93} & \textbf{80.57} & \textbf{81.26} \\
Adaptive & \textbf{82.38} & 80.63 & 80.01 & \textbf{73.44} & \textbf{78.48} & 81.33 & 81.87 & 80.00 & 80.43 \\
\end{tabular}
\label{table:summary}
\end{table}

We find that using a biased filter does not significantly increase the classification for any of the learning algorithms and that using a biased filter significantly decreases the classification accuracy for the LWL, na\"{i}ve Bayes, Ridor and RIPPER learning algorithms.
These results suggest that simply removing the misclassified instances by a single learning algorithm is not sufficient.
Bear in mind that these results reflect not adding any artificial noise to the training set.
In the case where artificial noise is added to the training set (as was commonly done in previous work), using a biased filter may result in an improvement in accuracy.
However, most real-world scenarios do not artificially add noise to their data set but are concerned with the inherent noise found within it.

For all of the learning algorithms, the ensemble filter significantly increases the classification accuracy over not filtering and over the other filtering techniques.
An ensemble generally provides better predictive performance than any of the constituent learning algorithms \cite{Polikar2006} and generally yields better results when the underlying ensembled models are diverse \cite{Kuncheva2003}.
Thus, by using a more powerful model, only the noisiest instances are removed.
This provides empirical evidence supporting the notion that filtering instances with low $p(\hat{y}_i|x_i)$ that are not dependent on a single hypothesis is preferred to filtering instances where the probability of the class is dependent on a particular hypothesis $p(\hat{y}_i|x_i,h)$ as outlined in Equation \ref{eq:sumOverH}.
% This also supports the findings by Smith et al. that class overlap is a primary contributor to an instance being misclassified \cite{Smith2012_IH}.

Surprisingly, the adaptive filter does not outperform the ensemble filter and in, one case, it does not even outperform training on unfiltered data.
Perhaps this is because it overfits the training data since the best accuracy is chosen on the training set.
Adaptive filtering has significantly better results when cross-validation is used to filter misclassified instances as opposed to removing misclassified instances that were also used to train the filtering algorithm.
Even with using the results with cross-validation, the results are not significantly better than using an ensemble filter.

\begin{table}
\caption{The \textit{p}-values from the Wilcoxon signed-ranks statistical significance test comparing not filtering with an ensemble filter.
The learning algorithms are ordered in descending order of \textit{p}-value from left to right.}
\setlength{\tabcolsep}{4.7pt}
\begin{tabular}{l|ccccccccc}
& RF& C4.5 & Rid & IB5 & NNge & MLP & LWL & RIP & NB\\
\hline
\textit{p}-val & 0.045 & 0.035 & 0.019 & 0.018 & 0.006 & 0.004 & 0.004 & $<0.001$ & $< 0.001$\\
\end{tabular}
\label{table:pVals}
\end{table}

Examining each learning algorithm individually, we find that some learning algorithms are more robust to noise than others.
To determine which learning algorithms are more robust to noise, we compare the accuracy of the learning algorithms without filtering to the accuracy obtained using an ensemble filter.
The \textit{p}-values from the Wilcoxon signed-ranks statistical significance test are shown in Table \ref{table:pVals} ordered from greatest (least significant impact) to least reading from left to right.
We see that random forests and decision trees are the most robust to noise as filtering has the least significant impact on their accuracy.
This is not too surprising given that the C4.5 algorithm was designed to take noise into account and random forests are built using decision trees.
Ridor and 5-nearest neighbor (IB5) are more robust to noise, but still greatly improve with filtering.
IB5 is more robust to noise since it compares with the 5 nearest neighbors of an instance.
If $K$ were set to 1, then filtering would have a greater effect on the accuracy.
Filtering has the most significant effect on the accuracy of the last five learning algorithms: MLP, NNge, LWL, RIPPER, and na\"{i}ve Bayes. 

\subsection{Analysis of When to Filter}
\label{section:filteringAnalysis}
Using only the inherent noise in a data set, the efficacy of filtering is limited and can be detrimental in some data sets.
Thus, we examine the cases in which filtering significantly improves the classification accuracy.
This investigation is similar to the recent work by S\'{a}ez et al. \cite{Saez2013} who investigate creating a set of rules to understand when to filter using a 1-nearest neighbor learning algorithm.
They use a set of data complexity measures from Ho and Basu \cite{Ho2002}.
The complexity measures are designed for binary classification problems, yet we do not limit ourselves to binary classification problems.
As such, we use a subset of the data complexity measures shown in Table \ref{table:complexityMeasures} that have been extended to handle multi-class problems \cite{DCoL}.
\begin{table}[t]
\caption{List of complexity measures from Ho and Basu \cite{Ho2002}.}
\label{table:complexityMeasures}
\begin{center}
\begin{tabular}{ l l}
%& Complexity Measure & Meta-feature\\
%& F1: Max Fisher's discriminant ratio &  \\
%& F3: Max individual feature efficiency & L1: Min sum of error distance by LP \\
F2:& \textbf{Volume of overlap region}:The overlap of the per-class bounding \\&boxes calculated for each attribute by normalizing the difference of \\&the maximum and minimum values from each class.\\
F3:& \textbf{Max individual feature efficiency}: For all of the features, the \\&maximum ratio of the number of instances not in the overlapping \\&region to the total number of instances.\\
F4:& \textbf{Collective feature efficiency}: F3 only return the ratio for the \\&attribute that maximizes the ratio. F4 is a measure for all of the \\&attributes. \\
N1:& \textbf{Fraction of points on class boundary}: The fraction of instances \\&in a data set that are connected to their nearest neighbors that have \\&a different class in a spanning tree. \\
N2:& \textbf{Ratio of ave intra/inter class NN dist}:  The average distance to \\&the nearest intra-class neighbors divided by the average distance to \\&the nearest inter-class neighbors.\\
N3:& \textbf{Error rate of 1NN classifier}: Leave-one-out error estimate of \\&1NN.\\
T1:& \textbf{Fraction of maximum covering spheres}: The normalized count \\&of the number of clusters of instances containing a single class \\
T2:& \textbf{Ave number of points per dimension}: Compares the number of \\&instances to the number of features.   \\
\end{tabular}
\end{center}
\end{table}
In addition, we also examine a set of hardness measures \cite{Smith2012_IH} shown in Table \ref{table:hardnessMeasures}.
The hardness measures are designed to determine and characterize instances that have a high likelihood of being misclassified.
We examine using the set of data complexity measures and the hardness measures to create rules and/or a classifier to determine when to use filtering.
We set up the classification problem similar to S\'{a}ez et al. where filtering is set to ``TRUE'' if filtering significantly improves the classification accuracy for a data set using the Wilcoxon signed-ranks test.
We also examine predicting the difference in accuracy between using and not using a filter.
Unlike S\'{a}ez et al., we find that the data complexity measures and the hardness measures do \textit{not} create a satisfactory classifier to determine when to filter.
Granted, we examine more learning algorithms and do not artificially add noise to the data sets which provides for few data sets where filtering significantly improves the classification accuracy.
In the study by S\'{a}ez et al., 75\% of the data sets had at least 5\% noise added providing more positive examples.
More future work is required to determine when to use filtering on unmodified data sets.
Based on our results, we would recommend always using an ensemble filter for all of the learning algorithms as it significantly outperforms the other filtering techniques.

\begin{table}[t]
\caption{List of hardness measures from Smith et al. \cite{Smith2012_IH}.}
\label{table:hardnessMeasures}
\begin{center}
\begin{tabular}{ll}
\textit{k}DN & \textbf{\textit{k}-Disagreeing Neighbors:} The percentage of the $k$ nearest \\&neighbors (using Euclidean distance) for an instance that do not \\&share its target class value.\\
DS & \textbf{Disjunct Size:} The number of instances in a disjunct divided by \\&the number of instances covered by the largest disjunct in a \\&data set in an unpruned decision tree inferred using C4.5 \cite{Quinlan1993}.\\
DCP & \textbf{Disjunct Class Percentage:} The number of instances in a \\&disjunct belonging to its class divided by the total number of \\&instances in the disjunct in a pruned decision tree.\\
TD & \textbf{Tree Depth:} The depth of the leaf node that classifies an \\&instance in an induced decision tree. \\
CL & \textbf{Class Likelihood:} The probability that an instance belongs to its \\&class given the input features. \\
CLD & \textbf{Class Likelihood Difference:} The difference between the class \\&likelihood of an instance and the maximum likelihood for all of the \\&other classes.\\
MV & \textbf{Minority Value:} The ratio of the number of instances sharing its \\&target class value to the number of instances in the majority class.\\
CB & \textbf{Class Balance:} The difference of the ratio of the number of \\&instances belonging to a class and the ratio of the classes if they \\&were distributed equally.\\
\end{tabular}
\end{center}
\end{table}

\subsection{Voting Ensemble VS. Filtering}
\label{section:filteringVsEnsemble}
In This section, we compare the results of filtering using an ensemble filter with a voting ensemble.
The voting ensemble uses the learning algorithms shown in Table \ref{table:LA} and the vote from each learning algorithm is equally weighted.
Table \ref{table:ensemble} compares the voting ensemble with using an ensemble filter on each of the investigated learning algorithms giving the average accuracy, the \textit{p}-value, and the number of times that the accuracy of a voting ensemble is greater than, equal to, or less than using an ensemble filter.
The results for each data set are provided in Table \ref{table:ensembleDS} in \ref{section:ensembleDS}.
With no artificially generated noise, a voting ensemble achieves significantly higher classification accuracy than an ensemble filter for each of the examined learning algorithms.
This is not too surprising considering that previous research has shown that ensemble methods address issues that are common to all non-ensemble learning algorithms \cite{Dietterich2000} and that ensemble methods generally obtain a greater accuracy than that from a single learning algorithm that makes up part of the ensemble \cite{Opitz1999}.
Considering the computational requirements for training, using a voting ensemble for classification rather than filtering appears to be more beneficial.

\begin{table}[t]
\caption{Summary of comparing a voting ensemble with filtering using an ensemble filter.
Using an ensemble filter significantly improves the classification accuracy over using an ensemble filter for all of the examined learning algorithms.}
\begin{tabular}{lc|ccccc}
& Ensemble & MLP & C4.5 & IB5 & LWL & NB \\\hline
 Acc & 84.37 & 83.40 & 81.61 & 80.85 & 73.48 & 78.92 \\\hline
 \multicolumn{2}{l|}{\textit{p}-value} & \textbf{0.008} & $\mathbf{<0.001}$ & $\mathbf{<0.001}$ & $\mathbf{<0.001}$ & $\mathbf{<0.001}$  \\
 \multicolumn{2}{l|}{$>,=,<$}& 33,1,20 & 43,1,10 & 42,2,10 & 47,1,6 & 41,0,13  \\\hline\\\hline
& Ensemble & NNge & RF & Rid & RIP \\\hline
 Acc & 84.37 & 81.59 & 82.93 & 80.57 & 80.76\\\hline
 \multicolumn{2}{l|}{\textit{p}-value} & $\mathbf{<0.001}$ & $\mathbf{<0.001}$ & $\mathbf{<0.001}$ & $\mathbf{<0.001}$ \\
 \multicolumn{2}{l|}{$>,=,<$}& 44,2,8 & 39,0,15 & 47,1,6 & 44,1,9 \\\hline
\end{tabular}
\label{table:ensemble}
\end{table}

\begin{table}
\caption{Comparison of a voting ensemble against using an ensemble filter on a subset of data sets where more than 10\% of the constituent instances are noisy .
The accuracy of the voting ensemble (``Ens'') is in bold if it is greater than the accuracies from using an ensemble filter for the investigated learning algorithms.
The accuracy from using an ensemble filter is in bold if it is higher than the accuracy from a voting ensemble.
The column ``Per'' refers to the percentage of instances in the data set that are considered noisy.}
\setlength{\tabcolsep}{1.1pt}
\begin{tabular}{lc|ccccccccc|c}
Data set& Ens & MLP & C4.5 & IB5 & LWL & NB & NNge & RF & Rid & RIP & Per\\\hline
breastc & 73.99 & 73.43 & \textbf{75.17} & \textbf{74.13} & 73.31 & 73.43 & \textbf{74.13} & 73.19 & \textbf{74.71} & \textbf{74.59} & 10.1 \\
arrhyth & \textbf{71.11} & 70.13 & 70.65 & 59.14 & 57.67 & 65.63 & 65.71 & 66.59 & 70.65 & 71.09 & 12.0 \\
contact & 76.67 & \textbf{83.33} & \textbf{83.33} & 76.39 & 76.39 & 76.39 & \textbf{80.56} & \textbf{80.56} & \textbf{79.17} & \textbf{77.78} & 12.5 \\
lungCan & 53.75 & 52.08 & \textbf{56.25} & 47.92 & \textbf{55.21} & \textbf{55.21} & \textbf{56.25} & 52.08 & 51.04 & \textbf{54.17} & 12.5 \\
yeast & \textbf{61.08} & 59.43 & 60.13 & 59.32 & 40.7 & 58.15 & 59.4 & \textbf{61.08} & 59.74 & 60.01 & 13.7 \\
cm1\_req & 75.73 & \textbf{76.40} & \textbf{77.53} & \textbf{77.53} & \textbf{76.78} & \textbf{77.53} & \textbf{77.15} & \textbf{76.78} & \textbf{77.53} & \textbf{76.78} & 16.9 \\
titanic & \textbf{78.72} & 78.66 & 78.68 & 78.59 & 77.9 & 77.77 & 78.68 & 78.68 & 78.28 & 78.68 & 16.9 \\
post-op & 69.78 & \textbf{71.11} & \textbf{71.11} & \textbf{71.11} & \textbf{71.11} & \textbf{71.11} & \textbf{71.11} & \textbf{71.11} & \textbf{71.11} & \textbf{71.11} & 26.7 \\
pri-tum & 48.08 & 47.79 & 41.2 & 45.82 & 34.42 & \textbf{48.57} & 44.44 & 45.23 & 39.82 & 40.41 & 32.2 \\
\hline\hline
Acc & 67.66 & \textbf{68.04} & \textbf{68.23} & 65.55 & 62.61 & 67.09 & 67.49 & 67.26 & 66.89 & 67.18\\\hline
\multicolumn{2}{l|}{\textit{p}-value} & 0.367 & 0.820 & 0.125 & 0.213 & 0.410 & 0.545 & 0.312 & 0.410 & 0.715  \\
\multicolumn{2}{l|}{$>,=<$}& 6,0,3 & 4,0,5 & 6,0,3 & 6,0,3 & 5,0,4 & 4,0,5 & 5,1,3 & 5,0,4 & 4,0,5 \\
\end{tabular}
\label{table:noisyDS}
\end{table}

Many previous studies \cite{Zhu2004,Lawrence2001,Brodley1999,Verbaeten2003} have shown that when a large amount of artificial noise is added to a data set (i.e. $\ge 10\%$), then filtering outperforms a voting ensemble.
We examine which of the 54 data sets have a high percentage of noise using instance hardness \cite{Smith2012_IH} to identify suspected noisy instances.
Instance hardness approximates the likelihood that an instance will be misclassified by evaluating the classification of an instance from a set of learning algorithms $\mathcal{L}$: $p(\hat{y}_i|x_i,\mathcal{L})$.
The set of learning algorithms $\mathcal{L}$ is composed of the learning algorithms shown in Table \ref{table:LA}.
The instances that have a probability greater than 0.9 of being misclassified we consider to be noisy instances.
Table \ref{table:noisyDS} shows the accuracies from a voting ensemble and the considered learning algorithms using an ensemble filter for the subset of data sets with more than 10\% noisy instances.
Examining the more noisy data sets shows that the gains from using an ensemble filter are more noticeable.
However, only 9 out of the 54 investigated data sets were identified as having more than 10\% noisy instances.
We ran a Wilcoxon signed-ranks test, but with the small sample size it is difficult to determine the statistical significance of using the ensemble filter over using a voting ensemble.
Based on the small sample provided here, training a learning algorithm on a filtered data set is statistically equivalent to training a voting ensemble classifier.
The computational complexity required to train an ensemble is less than that to train an ensemble for filtering followed by training another learning algorithm from the filtered data set.
A single learning algorithm trained on the filtered data set has the benefit that only one learning algorithm is queried for a novel instance.
Future work will include discovering if a smaller subset of learning algorithms for filtering approximates using the ensemble filter in order to reduce the computational complexity.

Examining the more noisy data sets shows that filtering has a more significant effect on classification accuracy, however, the amount of noise is not the only factor that needs to be considered.
For example, 32.2\% of the instances in the primary-tumor data set are noisy, yet only one learning algorithm achieves a greater classification accuracy than the voting ensemble.
On the other hand, the classification accuracy on the ar1 and ozone data sets for all of the considered learning algorithms trained on filtered data is greater than using a voting ensemble despite only having 3.3\% and 0.5\% noisy instances respectively.
Thus, there are other unknown data set features affecting when filtering is appropriate.
Future work also includes discovering and examining data set features that are indicative of when filtering should be used.

\begin{table}[t]
\caption{Comparison of a majority voting ensemble trained on unfiltered (Ens) and filtered data (FEns).
The value after ``FEns'' represents the percentage of learning algorithms that have to misclassify an instance for it to be filtered from the training set and ``Max'' uses the accuracy from the percentage that results in the greatest accuracy.
Training with unfiltered data is significantly better than training with filtered data for a voting ensemble.}
\center
\begin{tabular}{llc|cccc}
&& Ens & FEns 50 &  FEns 70 &  FEns 90 &  FEns Max \\\hline
\multirow{3}{*}{\begin{sideways}All\end{sideways}} &Accuracy & 84.37 & 83.40 & 82.21 & 73.96 & 83.62\\
& \multicolumn{2}{l|}{\textit{p}-value} & $\mathbf{<0.001}$ & $\mathbf{<0.001}$ & $\mathbf{<0.001}$ & $\mathbf{<0.001}$ \\
& \multicolumn{2}{l|}{greater-equal-less} & 42,3,9 & 44,2,8 & 48,1,5 & 39,2,13 \\\hline
\multirow{3}{*}{\begin{sideways}Noisy\end{sideways}} & Accuracy & 67.66 & 67.00 & 67.47 & 60.52 & 67.93\\
 & \multicolumn{2}{l|}{\textit{p}-value} & 0.102 & 0.455 & \textbf{0.049} & 0.633 \\
 & \multicolumn{2}{l|}{greater-equal-less} & 7,0,2 & 5,0,4 & 6,0,3 & 5,0,4 \\\hline
\multirow{3}{*}{\begin{sideways}$<90\%$\end{sideways}}  & Accuracy & 78.49 & 77.19 & 75.74 & 66.02 & 77.44\\
& p-values & & $\mathbf{<0.001}$ & $\mathbf{<0.001}$ & $\mathbf{<0.001}$ & $\mathbf{<0.001}$ \\
 & greater-equal-less & & 31,0,6 & 30,1,6 & 34,0,3 & 28,0,9 \\\hline
 \multirow{3}{*}{\begin{sideways}$<80\%$\end{sideways}} & Accuracy & 74.70 & 73.08 & 71.41 & 61.49 & 73.41\\
 & p-values &  & $\mathbf{<0.001}$ & $\mathbf{<0.001}$ & $\mathbf{<0.001}$ & \textbf{0.002} \\
 & greater-equal-less &  & 24,0,3 & 22,0,5 & 24,0,3 & 21,0,6 \\\hline
\multirow{3}{*}{\begin{sideways}$<70\%$\end{sideways}} & Accuracy & 64.65 & 61.99 & 60.41 & 51.04 & 62.25\\
 & p-values & & \textbf{0.001} & \textbf{0.002} & $\mathbf{<0.001}$ & \textbf{0.009} \\
 & greater-equal-less & & 10,0,1 & 10,0,1 & 10,0,1 & 10,0,1 \\\hline
\multirow{3}{*}{\begin{sideways}$<60\%$\end{sideways}} & Accuracy & 58.44 & 55.81 & 53.49 & 42.56 & 56.12\\
 & p-values & & \textbf{0.016} & \textbf{0.016} & \textbf{0.016} & \textbf{0.016} \\
 & greater-equal-less & & 6,0,0 & 6,0,0 & 6,0,0 & 6,0,0 \\\hline
\multirow{3}{*}{\begin{sideways}$<50\%$\end{sideways}} & Accuracy & 50.92 & 48.22 & 48.07 & 38.61 & 49.16\\
& p-values & & 0.250 & 0.250 & 0.250 & 0.250 \\
 & greater-equal-less & & 2,0,0 & 2,0,0 & 2,0,0 & 2,0,0 \\\hline

\end{tabular}
\label{table:ensVsFilteredEns}
\end{table}

We further investigate the robustness of the majority voting ensemble to noise by applying an ensemble filter to the training data for the voting ensemble.
We find that a majority voting ensemble is significantly better \textit{without} filtering.
The summary results are shown in Table \ref{table:ensVsFilteredEns} and the full results for each data set can be found in Table B.18 in \ref{section:ensembleDS}.
Table \ref{table:ensVsFilteredEns} divides the data sets into subsets that have more than 10\% noisy instances (``Noisy''), and those that have an original accuracy less than 90\%, 80\%, 70\%, 60\%, and 50\% averaged across the investigated learning algorithms ($< N\%$).
Even with harder data sets and more noisy instances, using unfiltered training data produces significantly higher classification accuracy for the voting ensemble.
Thus, we find that a majority voting ensemble is more robust to noise than filtering in most cases.
The strength of a voting ensemble comes from the diversity of the ensembled learning algorithms.
However, the inferred models from the learning algorithms trained on the filtered training data are less diverse since the diversity often comes from how a learning algorithm treats a noisy instance, lessening the power of the voting ensemble.
This is evidenced as we examined a voting ensemble consisting of C4.5, random forest, and Ridor which are three of the more similar learning algorithms using unsupervised meta-learning (see Section \ref{section:methodology}).
When trained on the filtered training data, the less diverse voting ensemble achieves a significantly lower classification average accuracy of 82.09\% compared to 83.62\% from the voting ensemble composed of the 9 examined learning algorithms.

\section{Conclusions}
\label{section:conclusions}
In this paper, we presented an extensive empirical evaluation of misclassification filters on a set of 54 data sets and 9 diverse learning algorithms.
As opposed to other work on filtering, we used a large set of data sets and learning algorithms and we did not artificially add noise to the data set.
In previous work, noise was added to a data set to verify that the noise filtering method was effective and that filtering was more effective when more noise was present.
However, the artificial noise may not be representative of the actual noise and the impact of filtering on an unmodified data set is not always clear.

We examined each learning algorithm individually as a filter as well as using all of the learning algorithms combined as an ensemble filter.
We also presented an adaptive filtering algorithm that greedily searches the set of candidate learning algorithms for filtering for a specific data set and learning algorithm combination.
We found that, without artificially adding label noise, using the same learning algorithm for filtering and for inferring a model of the data can be significantly detrimental and does not significantly increase the classification accuracy even when examining harder data sets.
We also examined using a set of data set features to induce rules that indicate when to use filtering, but did not find a set of rules that significantly improved the results.
Using an ensemble filter significantly improved the accuracy over not filtering and outperformed both the adaptive filtering method and using each learning algorithm individually as a filter for all of the investigated learning algorithms.

We also compared filtering with a voting ensemble and found that a voting ensemble achieves significantly higher classification accuracy than any of the other considered learning algorithms trained on filtered data.
A majority voting ensemble trained on unfiltered data significantly outperforms a voting ensemble trained on filtered data.
Thus, a voting ensemble exhibits robustness to noise in the training set and is preferable to filtering.

\bibliographystyle{elsarticle-num}
\bibliography {../../../bibliography}

\begin{thebibliography}{10}
\expandafter\ifx\csname url\endcsname\relax
  \def\url#1{\texttt{#1}}\fi
\expandafter\ifx\csname urlprefix\endcsname\relax\def\urlprefix{URL }\fi
\expandafter\ifx\csname href\endcsname\relax
  \def\href#1#2{#2} \def\path#1{#1}\fi

\bibitem{Zhu2004}
X.~Zhu, X.~Wu, Class noise vs. attribute noise: a quantitative study of their
  impacts, Artificial Intelligence Review 22 (2004) 177--210.

\bibitem{Quinlan1993}
J.~R. Quinlan, C4.5: Programs for Machine Learning, Morgan Kaufmann, San Mateo,
  CA, USA, 1993.

\bibitem{Wilson1972}
D.~L. Wilson, Asymptotic properties of nearest neighbor rules using edited
  data, IEEE Transactions on Systems, Man, and Cybernetics~(2-3) (1972)
  408--421.

\bibitem{Brodley1999}
C.~E. Brodley, M.~A. Friedl, Identifying mislabeled training data, Journal of
  Artificial Intelligence Research 11 (1999) 131--167.

\bibitem{Teng2003}
C.-M. Teng, Combining noise correction with feature selection, in:
  Y.~Kambayashi, M.~K. Mohania, W.~W{\"o}{\ss} (Eds.), DaWaK, Vol. 2737 of
  Lecture Notes in Computer Science, Springer, 2003, pp. 340--349.

\bibitem{John95}
G.~H. John, Robust decision trees: Removing outliers from databases, in:
  Knowledge Discovery and Data Mining, 1995, pp. 174--179.

\bibitem{Tomek1976}
I.~Tomek, An experiment with the edited nearest-neighbor rule, IEEE
  Transactions on Systems, Man, and Cybernetics 6 (1976) 448--452.

\bibitem{Wilson2000}
D.~R. Wilson, T.~R. Martinez, Reduction techniques for instance-based learning
  algorithms, Machine Learning 38~(3) (2000) 257--286.

\bibitem{bishop2006pattern}
C.~M. Bishop, N.~M. Nasrabadi, Pattern Recognition and Machine Learning,
  Vol.~1, springer New York, 2006.

\bibitem{Schapire1990}
R.~E. Schapire, The strength of weak learnability, Machine Learning 5 (1990)
  197--227.

\bibitem{Freund1995}
Y.~Freund, Boosting a weak learning algorithm by majority, in: Proceedings of
  the Third Annual Workshop on Computational Learning Theory, 1990, pp.
  202--216.

\bibitem{Servedio2003}
R.~A. Servedio, Smooth boosting and learning with malicious noise, Journal of
  Machine Learning Research 4 (2003) 633--648.

\bibitem{Collobert2006}
R.~Collobert, F.~Sinz, J.~Weston, L.~Bottou, Trading convexity for scalability,
  in: Proceedings of the 23rd International Conference on Machine learning,
  2006, pp. 201--208.

\bibitem{Lawrence2001}
N.~D. Lawrence, B.~Sch{\"o}lkopf, Estimating a kernel fisher discriminant in
  the presence of label noise, in: In Proceedings of the 18th International
  Conference on Machine Learning, 2001, pp. 306--313.

\bibitem{Gamberger2000}
D.~Gamberger, N.~Lavra\v{c}, S.~D\v{z}eroski, Noise detection and elimination
  in data preprocessing: Experiments in medical domains, Applied Artificial
  Intelligence 14~(2) (2000) 205--223.

\bibitem{Smith2011}
M.~R. Smith, T.~Martinez, Improving classification accuracy by identifying and
  removing instances that should be misclassified, in: Proceedings of the IEEE
  International Joint Conference on Neural Networks, 2011, pp. 2690--2697.

\bibitem{Verbaeten2003}
S.~Verbaeten, A.~{Van Assche}, Ensemble methods for noise elimination in
  classification problems, in: Proceedings of the 4th international conference
  on Multiple classifier systems, MCS'03, Springer-Verlag, Berlin, Heidelberg,
  2003, pp. 317--325.

\bibitem{Segata2009}
N.~Segata, E.~Blanzieri, P.~Cunningham, A scalable noise reduction technique
  for large case-based systems, in: Proceedings of the 8th International
  Conference on Case-Based Reasoning: Case-Based Reasoning Research and
  Development, 2009, pp. 328--342.

\bibitem{Xinchuan2003}
X.~Zeng, T.~R. Martinez, A noise filtering method using neural networks, in:
  Proc. of the int. Workshop of Soft Comput. Techniques in Instrumentation,
  Measurement and Related Applications, 2003.

\bibitem{Rebbapragada2007}
U.~Rebbapragada, C.~Brodley, Class noise mitigation through instance weighting,
  in: Machine Learning: ECML 2007, Vol. 4701 of Lecture Notes in Computer
  Science, Springer Berlin Heidelberg, 2007, pp. 708--715.

\bibitem{Smith_RIDL}
M.~R. Smith, T.~Martinez, Reducing the effects of detrimental instances, in:
  Submission, 2013.

\bibitem{zeng.ida2001}
X.~Zeng, T.~R. Martinez, An algorithm for correcting mislabeled data,
  Intelligent Data Analysis 5 (2001) 491--502.

\bibitem{Teng2000}
C.-M. Teng, Evaluating noise correction, in: PRICAI, 2000, pp. 188--198.

\bibitem{Ng+Jordan:2001}
A.~Y. Ng, M.~I. Jordan, On discriminative vs. generative classifiers: A
  comparison of logistic regression and naive bayes, in: Advances in Neural
  Information Processing Systems 14, 2001, pp. 841--848.

\bibitem{Lee2011}
J.~Lee, C.~Giraud-Carrier, A metric for unsupervised metalearning, Intelligent
  Data Analysis 15~(6) (2011) 827--841.

\bibitem{Peterson2005}
A.~H. Peterson, T.~R. Martinez, Estimating the potential for combining learning
  models, in: Proceedings of the ICML Workshop on Meta-Learning, 2005, pp.
  68--75.

\bibitem{weka2009}
M.~Hall, E.~Frank, G.~Holmes, B.~Pfahringer, P.~Reutemann, I.~H. Witten, The
  weka data mining software: an update, SIGKDD Explorations Newsletter 11~(1)
  (2009) 10--18.

\bibitem{Thomson1996}
K.~Thomson, R.~J. McQueen, Machine learning applied to fourteen agricultural
  datasets, Tech. Rep. 96/18, The University of Waikato (September 1996).

\bibitem{Salojärvi05}
J.~Saloj{\"a}rvi, K.~Puolam{\"a}ki, J.~Simola, L.~Kovanen, I.~Kojo, S.~Kaski,
  Inferring relevance from eye movements: Feature extraction, Tech. Rep. A82,
  Helsinki University of Technology (March 2005).

\bibitem{promise}
J.~{Sayyad Shirabad}, T.~Menzies,
  \href{http://promise.site.uottawa.ca/SERepository/}{{The {PROMISE} Repository
  of Software Engineering Databases.}}, School of Information Technology and
  Engineering, University of Ottawa, Canada (2005).
\newline\urlprefix\url{http://promise.site.uottawa.ca/SERepository/}

\bibitem{gemlrep}
G.~Stiglic, P.~Kokol, \href{http://gemler.fzv.uni-mb.si/}{{GEML}er: Gene
  expression machine learning repository}, University of Maribor, Faculty of
  Health Sciences (2009).
\newline\urlprefix\url{http://gemler.fzv.uni-mb.si/}

\bibitem{Demsar2006}
J.~Dem\v{s}ar, Statistical comparisons of classifiers over multiple data sets,
  Journal of Machine Learning Research 7.

\bibitem{Polikar2006}
R.~Polikar, Ensemble based systems in decision making, IEEE Circuits and
  Systems Magazine 6~(3) (2006) 21--45.

\bibitem{Kuncheva2003}
L.~I. Kuncheva, C.~J. Whitaker, Measures of diversity in classifier ensembles
  and their relationship with the ensemble accuracy., Machine Learning 51~(2)
  (2003) 181--207.

\bibitem{Saez2013}
J.~A. S{\'a}ez, J.~Luengo, F.~Herrera, Predicting noise filtering efficacy with
  data complexity measures for nearest neighbor classification, Pattern
  Recognition 46~(1) (2013) 355--364.

\bibitem{Ho2002}
T.~K. Ho, M.~Basu, Complexity measures of supervised classification problems,
  IEEE Transactions on Pattern Analysis and Machine Intelligence 24 (2002)
  289--300.

\bibitem{DCoL}
A.~Orriols-Puig, N.~Maci{\`a}, E.~Bernad{\'o}-Mansilla, T.~K. Ho, Documentation
  for the data complexity library in c++, Tech. Rep. 2009001, La Salle -
  Universitat Ramon Llull (April 2009).

\bibitem{Smith2012_IH}
M.~R. Smith, T.~Martinez, C.~Giraud-Carrier, An instance level analysis of data
  complexity, Machine Learning (2013) In press\href
  {http://dx.doi.org/10.1007/s10994-013-5422-z}
  {\path{doi:10.1007/s10994-013-5422-z}}.

\bibitem{Dietterich2000}
T.~G. Dietterich, Ensemble methods in machine learning, in: Multiple Classifier
  Systems, Vol. 1857 of Lecture Notes in Computer Science, Springer, 2000, pp.
  1--15.

\bibitem{Opitz1999}
D.~W. Opitz, R.~Maclin, Popular ensemble methods: An empirical study., Journal
  of Artificial Intelligence Research 11 (1999) 169--198.

\end{thebibliography}

\appendix
\section{Statistical Significance Tables}
\label{section:sigTests}
This section provides the results from the statistical significance tests comparing not filtering with filtering with a biased filter, an ensemble filter, and the adaptive filter for the investigated learning algorithms.
The results are in Tables \ref{table:sigBackprop} - \ref{table:sigRIPPER}.
The \textit{p}-values with a value less than 0.05 are in bold and ``greater-equal-less'' refers to the number of times that the algorithm listed in the row is greater than, equal to, or less than the algorithm listed in the column.

\begin{table}[hb]
\caption{Pair-wise comparison of filtering for multilayer perceptrons trained with backpropagation.}
\begin{tabular}{ll|cccc}
% && Orig & Biased & Ensemble & Adaptive \\\hline
%  & Accuracy & 81.23 & 81.72 & \textbf{83.40} & 82.38\\\hline
% \multirow{2}{*}{Orig} & p-values & 1 & 0.709 & 1 & 0.987 \\
%  & greater-equal-less & 0,54,0 & 26,1,27 & 9,1,44 & 20,1,33 \\\hline
% \multirow{2}{*}{Biased} & p-values & 0.295 & 1 & 1 & 0.985 \\
%  & greater-equal-less & 27,1,26 & 0,54,0 & 10,5,39 & 17,3,34 \\\hline
% \multirow{2}{*}{Ensemble} & p-values & $\mathbf{<0.001}$ & $\mathbf{<0.001}$ & 1 & $\mathbf{<0.001}$ \\
%  & greater-equal-less & 44,1,9 & 39,5,10 & 0,54,0 & 38,3,13 \\\hline
% \multirow{2}{*}{Adaptive} & p-values & \textbf{0.014} & \textbf{0.015} & 1 & 1 \\
%  & greater-equal-less & 33,1,20 & 34,3,17 & 13,3,38 & 0,54,0 \\\hline
&& Orig & Biased & Ensemble & Greedy \\\hline
 & Accuracy & 81.74 & 81.87 & 83.33 & 82.33\\\hline
\multirow{2}{*}{Orig} & p-values & 1 & 0.771 & 1.000 & 0.953 \\
 & greater-equal-less & 0,54,0 & 25,2,27 & 16,2,36 & 18,3,33 \\\hline
\multirow{2}{*}{Biased} & p-values & 0.232 & 1 & 1 & 0.957 \\
 & greater-equal-less & 27,2,25 & 0,54,0 & 9,4,41 & 23,2,29 \\\hline
\multirow{2}{*}{Ensemble} & p-values & $\mathbf{<0.001}$ & $\mathbf{<0.001}$ & 1 & $\mathbf{<0.001}$ \\
 & greater-equal-less & 36,2,16 & 41,4,9 & 0,54,0 & 40,1,13 \\\hline
\multirow{2}{*}{Greedy} & p-values & \textbf{0.048} & \textbf{0.044} & 1 & 1 \\
 & greater-equal-less & 33,3,18 & 29,2,23 & 13,1,40 & 0,54,0 \\\hline
\end{tabular}
\label{table:sigBackprop}
\end{table}

\begin{table}
\caption{Pair-wise comparison of filtering for decision trees.}
\begin{tabular}{ll|cccc}
 & Accuracy & 80.80 & 80.83 & 81.59 & 80.56\\\hline
\multirow{2}{*}{Orig} & p-values & 1 & 0.460 & 1.000 & 0.221 \\
 & greater-equal-less & 0,54,0 & 26,5,23 & 17,3,34 & 29,2,23 \\\hline
\multirow{2}{*}{Biased} & p-values & 0.544 & 1 & 0.999 & 0.271 \\
 & greater-equal-less & 23,5,26 & 0,54,0 & 17,5,32 & 29,1,24 \\\hline
\multirow{2}{*}{Ensemble} & p-values & $\mathbf{<0.001}$ & \textbf{0.001} & 1 & $\mathbf{<0.001}$ \\
 & greater-equal-less & 34,3,17 & 32,5,17 & 0,54,0 & 44,2,8 \\\hline
\multirow{2}{*}{Greedy} & p-values & 0.782 & 0.732 & 1 & 1 \\
 & greater-equal-less & 23,2,29 & 24,1,29 & 8,2,44 & 0,54,0 \\\hline
\end{tabular}
\label{table:sigC4.5}
\end{table}

\begin{table}
\caption{Pair-wise comparison of filtering for 5-nearest neighbors.}
\begin{tabular}{ll|cccc}
&& Orig & Biased & Ensemble & Greedy \\\hline
 & Accuracy & 79.91 & 79.40 & 80.83 & 79.91\\\hline
\multirow{2}{*}{Orig} & p-values & 1 & 0.693 & 0.985 & 0.877 \\
 & greater-equal-less & 0,54,0 & 25,1,28 & 17,2,35 & 20,2,32 \\\hline
\multirow{2}{*}{Biased} & p-values & 0.310 & 1 & 1 & 0.999 \\
 & greater-equal-less & 28,1,25 & 0,54,0 & 5,4,45 & 17,1,36 \\\hline
\multirow{2}{*}{Ensemble} & p-values & \textbf{0.015} & $\mathbf{<0.001}$ & 1 & $\mathbf{<0.001}$ \\
 & greater-equal-less & 35,2,17 & 45,4,5 & 0,54,0 & 44,1,9 \\\hline
\multirow{2}{*}{Greedy} & p-values & 0.125 & \textbf{0.001} & 1 & 1 \\
 & greater-equal-less & 32,2,20 & 36,1,17 & 9,1,44 & 0,54,0 \\\hline
\end{tabular}
\label{table:sig5NN}
\end{table}

\begin{table}
\caption{Pair-wise comparison of filtering for locally weighted learning (LWL).}
\begin{tabular}{ll|cccc}
&& Orig & Biased & Ensemble & Greedy \\\hline
 & Accuracy & 72.80 & 70.91 & 73.48 & 73.44\\\hline
\multirow{2}{*}{Orig} & p-values & 1 & $\mathbf{<0.001}$ & 0.992 & 0.988 \\
 & greater-equal-less & 0,54,0 & 34,11,9 & 14,9,31 & 16,8,30 \\\hline
\multirow{2}{*}{Biased} & p-values & 0.999 & 1 & 1 & 1 \\
 & greater-equal-less & 9,11,34 & 0,54,0 & 3,12,39 & 9,10,35 \\\hline
\multirow{2}{*}{Ensemble} & p-values & \textbf{0.009} & $\mathbf{<0.001}$ & 1 & 0.595 \\
 & greater-equal-less & 31,9,14 & 39,12,3 & 0,54,0 & 19,8,27 \\\hline
\multirow{2}{*}{Greedy} & p-values & \textbf{0.013} & $\mathbf{<0.001}$ & 0.409 & 1 \\
 & greater-equal-less & 30,8,16 & 35,10,9 & 27,8,19 & 0,54,0 \\\hline
\end{tabular}
\label{table:sigLWL}
\end{table}

\begin{table}
\caption{Pair-wise comparison of filtering for na\"{i}ve Bayes.}
\begin{tabular}{ll|cccc}
&& Orig & Biased & Ensemble & Greedy \\\hline
 & Accuracy & 76.94 & 75.84 & 78.82 & 78.45\\\hline
\multirow{2}{*}{Orig} & p-values & 1 & \textbf{0.001} & 1.000 & 0.985 \\
 & greater-equal-less & 0,54,0 & 38,0,16 & 17,4,33 & 24,1,29 \\\hline
\multirow{2}{*}{Biased} & p-values & 0.999 & 1 & 1 & 1 \\
 & greater-equal-less & 16,0,38 & 0,54,0 & 4,2,48 & 10,2,42 \\\hline
\multirow{2}{*}{Ensemble} & p-values & $\mathbf{<0.001}$ & $\mathbf{<0.001}$ & 1 & \textbf{0.012} \\
 & greater-equal-less & 33,4,17 & 48,2,4 & 0,54,0 & 32,4,18 \\\hline
\multirow{2}{*}{Greedy} & p-values & \textbf{0.016} & $\mathbf{<0.001}$ & 0.988 & 1 \\
 & greater-equal-less & 29,1,24 & 42,2,10 & 18,4,32 & 0,54,0 \\\hline
\end{tabular}
\label{table:sigNB}
\end{table}

\begin{table}
\caption{Pair-wise comparison of filtering for NNge.}
\begin{tabular}{ll|cccc}
&& Orig & Biased & Ensemble & Greedy \\\hline
 & Accuracy & 80.62 & 80.30 & 82.18 & 81.32\\\hline
\multirow{2}{*}{Orig} & p-values & 1 & 0.080 & 1.000 & 0.888 \\
 & greater-equal-less & 0,52,0 & 25,5,22 & 12,4,36 & 24,1,27 \\\hline
\multirow{2}{*}{Biased} & p-values & 0.921 & 1 & 1 & 0.992 \\
 & greater-equal-less & 22,5,25 & 0,52,0 & 12,2,38 & 18,2,32 \\\hline
\multirow{2}{*}{Ensemble} & p-values & $\mathbf{<0.001}$ & $\mathbf{<0.001}$ & 1 & $\mathbf{<0.001}$ \\
 & greater-equal-less & 36,4,12 & 38,2,12 & 0,52,0 & 41,2,9 \\\hline
\multirow{2}{*}{Greedy} & p-values & 0.114 & \textbf{0.008} & 1 & 1 \\
 & greater-equal-less & 27,1,24 & 32,2,18 & 9,2,41 & 0,52,0 \\\hline
\end{tabular}
\label{table:sigNNge}
\end{table}

\begin{table}
\caption{Pair-wise comparison of filtering for random forests.}
\begin{tabular}{ll|cccc}
&& Orig & Biased & Ensemble & Greedy \\\hline
 & Accuracy & 82.28 & 82.21 & 82.92 & 81.85\\\hline
\multirow{2}{*}{Orig} & p-values & 1 & 0.408 & 0.981 & \textbf{0.022} \\
 & greater-equal-less & 0,54,0 & 28,2,24 & 23,1,30 & 35,2,17 \\\hline
\multirow{2}{*}{Biased} & p-values & 0.595 & 1 & 0.992 & 0.084 \\
 & greater-equal-less & 24,2,28 & 0,54,0 & 22,4,28 & 31,2,21 \\\hline
\multirow{2}{*}{Ensemble} & p-values & \textbf{0.020} & \textbf{0.009} & 1 & $\mathbf{<0.001}$ \\
 & greater-equal-less & 30,1,23 & 28,4,22 & 0,54,0 & 46,1,7 \\\hline
\multirow{2}{*}{Greedy} & p-values & 0.979 & 0.918 & 1 & 1 \\
 & greater-equal-less & 17,2,35 & 21,2,31 & 7,1,46 & 0,54,0 \\\hline
\end{tabular}
\label{table:sigRandForest}
\end{table}

\begin{table}
\caption{Pair-wise comparison of filtering for Ridor.}
\begin{tabular}{ll|cccc}
&& Orig & Biased & Ensemble & Greedy \\\hline
 & Accuracy & 79.90 & 79.16 & 80.56 & 79.96\\\hline
\multirow{2}{*}{Orig} & p-values & 1 & \textbf{0.016} & 1.000 & 0.895 \\
 & greater-equal-less & 0,54,0 & 33,2,19 & 15,1,38 & 20,3,31 \\\hline
\multirow{2}{*}{Biased} & p-values & 0.985 & 1 & 1 & 0.998 \\
 & greater-equal-less & 19,2,33 & 0,54,0 & 7,3,44 & 17,2,35 \\\hline
\multirow{2}{*}{Ensemble} & p-values & $\mathbf{<0.001}$ & $\mathbf{<0.001}$ & 1 & $\mathbf{<0.001}$ \\
 & greater-equal-less & 38,1,15 & 44,3,7 & 0,54,0 & 36,3,15 \\\hline
\multirow{2}{*}{Greedy} & p-values & 0.107 & \textbf{0.002} & 1.000 & 1 \\
 & greater-equal-less & 31,3,20 & 35,2,17 & 15,3,36 & 0,54,0 \\\hline

\end{tabular}
\label{table:sigRidor}
\end{table}

\begin{table}
\caption{Pair-wise comparison of filtering for RIPPER.}
\begin{tabular}{ll|cccc}
&& Orig & Biased & Ensemble & Greedy \\\hline
 & Accuracy & 80.34 & 79.98 & 81.25 & 80.41\\\hline
\multirow{2}{*}{Orig} & p-values & 1 & \textbf{0.040} & 1 & 0.704 \\
 & greater-equal-less & 0,53,0 & 30,2,21 & 11,2,40 & 21,6,26 \\\hline
\multirow{2}{*}{Biased} & p-values & 0.961 & 1 & 1 & 0.989 \\
 & greater-equal-less & 21,2,30 & 0,53,0 & 8,1,44 & 19,4,30 \\\hline
\multirow{2}{*}{Ensemble} & p-values & $\mathbf{<0.001}$ & $\mathbf{<0.001}$ & 1 & $\mathbf{<0.001}$ \\
 & greater-equal-less & 40,2,11 & 44,1,8 & 0,53,0 & 38,4,11 \\\hline
\multirow{2}{*}{Greedy} & p-values & 0.300 & \textbf{0.011} & 1 & 1 \\
 & greater-equal-less & 26,6,21 & 30,4,19 & 11,4,38 & 0,53,0 \\\hline
\end{tabular}
\label{table:sigRIPPER}
\end{table}
\newpage
\section{Ensemble Results for Each Data Set}
\label{section:ensembleDS}
This section provides the results for each data set comparing a voting ensemble with filtering using an ensemble filter for each investigated learning algorithm as well as filtering using an ensemble filter for a voting ensemble.
The results comparing a voting ensemble with filtering for each investigated non-ensembled learning algorithm are shown in Table \ref{table:ensembleDS}.
The bold values represent the highest classification accuracy and the rows highlighted in gray are the data sets where filtering with an ensemble filter increased the accuracy over the voting ensemble for all learning algorithms.
The results comparing a voting ensemble with a filtered voting ensemble are shown in Table \ref{table:ensVsFilteredEnsDSs}.
The bold values for the ``Ens'' column represent if the voting ensemble trained on unfiltered data achieves higher accuracy while the bold values for the ``FEns'' columns represent if the voting ensemble trained on filtered data achieves higher accuracy than the voting ensemble trained on unfiltered data.

\begin{table}
\caption{Comparison of the accuracy for each data set using a voting ensemble (Ens) with using an ensemble filter for the investigated learning algorithms.
The column ``Per'' refers to the percentage of instances that have a $p(\hat{y}_i|x_i)$ greater than or equal to 90\%.
The rows in gray represent those datasets where filtering with an ensemble filter increased the accuracy over the voting ensemble for all learning algorithms.}
\setlength{\tabcolsep}{0.8pt}
\begin{tabular}{l|ccccccccccc}
& Ens & MLP & C4.5 & IB5 & LWL & NB & NNge & RF & Rid & RIP & Per\\
\hline
anneal & 98.08 & \textbf{98.29} & 91.72 & 92.91 & 92.72 & 83.93 & 92.87 & 94.8 & 96.59 & 94.84 & 0.33 \\
AP-BU & 97.61 & 96.87 & 94.87 & 96.87 & 93.3 & 96.72 & 96.72 & \textbf{98.01} & 93.59 & 94.73 & 0.85 \\
\rowcolor{gray} ar1 & 90.08 & 92.29 & \textbf{92.56} & \textbf{92.56} & 92.29 & 92.29 & 92.29 & 92.29 & \textbf{92.56} & \textbf{92.56} & 3.31 \\
arrhyth & \textbf{71.11} & 70.13 & 70.65 & 59.14 & 57.67 & 65.63 & 65.71 & 66.59 & 70.65 & 71.09 & \textbf{11.95} \\
audiolo & \textbf{78.94} & 78.61 & 76.99 & 62.54 & 47.05 & 73.01 & 72.42 & 73.6 & 71.24 & 73.89 & 7.08 \\
autos & \textbf{83.51} & 78.54 & 79.84 & 64.72 & 51.71 & 56.1 & 74.8 & 82.6 & 69.59 & 76.1 & 4.88 \\
badges2 & \textbf{100} & \textbf{100} & \textbf{100} & \textbf{100} & \textbf{100} & 99.66 & \textbf{100} & 99.89 & \textbf{100} & \textbf{100} & 0.00 \\
balance & 88.45 & \textbf{90.35} & 78.67 & 89.65 & 60.59 & 89.97 & 82.56 & 82.77 & 79.68 & 79.09 & 4.16 \\
breastc & 73.99 & 73.43 & \textbf{75.17} & 74.13 & 73.31 & 73.43 & 74.13 & 73.19 & 74.71 & 74.59 & \textbf{10.14} \\
breastw & 96.88 & \textbf{97.00} & 95.14 & 96.76 & 92.61 & 95.95 & 95.99 & 96.57 & 95.61 & 95.8 & 1.72 \\
bupa & 71.3 & \textbf{71.5} & 66.47 & 62.71 & 60.29 & 59.03 & 65.31 & 69.28 & 67.44 & 68.02 & 2.61 \\
carEval & 96.7 & \textbf{98.82} & 92.09 & 92.77 & 70.02 & 85.22 & 94.21 & 92.46 & 95.72 & 87.15 & 0.00 \\
chess & \textbf{99.53} & 99.41 & 99.44 & 96.17 & 72.15 & 87.85 & 98.56 & 98.77 & 98.72 & 99.21 & 0.03 \\
\rowcolor{gray} cm1\_req & 75.73 & 76.4 & \textbf{77.53} & \textbf{77.53} & 76.78 & \textbf{77.53} & 77.15 & 76.78 & \textbf{77.53} & 76.78 & \textbf{16.85} \\
colic & 85.33 & \textbf{86.41} & 85.78 & 82.97 & 81.52 & 83.33 & 84.42 & 85.69 & 84.42 & 85.96 & 4.35 \\
contact & 76.67 & \textbf{83.33} & \textbf{83.33} & 76.39 & 76.39 & 76.39 & 80.56 & 80.56 & 79.17 & 77.78 & \textbf{12.50} \\
credita & \textbf{86.64} & 85.7 & 85.99 & 86.62 & 85.51 & 81.64 & 85.6 & 86.04 & 85.85 & 86.09 & 4.35 \\
creditg & \textbf{75.64} & 75.07 & 73.17 & 73.37 & 70.03 & 74.8 & 73.33 & 74.2 & 71.97 & 72.6 & 5.20 \\
derma & \textbf{97.43} & 97.09 & 93.99 & 96.08 & 87.61 & 97.36 & 95.36 & 95.99 & 94.35 & 88.8 & 0.00 \\
desh & 74.32 & 70.78 & 69.55 & 65.84 & 71.6 & 62.14 & 67.49 & \textbf{74.49} & 71.6 & 71.6 & 7.41 \\
ecoli & \textbf{87.44} & 86.21 & 84.72 & 87.2 & 65.87 & 86.9 & 85.22 & 85.71 & 83.73 & 82.74 & 4.76 \\
eucalyp & \textbf{65.11} & 63.32 & 62.86 & 55.8 & 51.04 & 57.52 & 56.84 & 56.88 & 61.73 & 63.32 & 6.66 \\
eye-mov & \textbf{64.76} & 54.34 & 63.72 & 54.93 & 42.88 & 44.11 & 48.13 & 62.8 & 54.11 & 56.04 & 1.77 \\
glass & 74.02 & 66.04 & 68.07 & 66.51 & 52.18 & 53.58 & 71.5 & \textbf{74.92} & 68.85 & 68.85 & 5.14 \\
heart-c & 83.83 & 83.39 & 77.56 & 83.17 & 75.69 & \textbf{83.94} & 79.98 & 81.63 & 79.65 & 80.97 & 3.30 \\
heart-h & 82.93 & 83.22 & 81.63 & \textbf{84.69} & 80.16 & 84.47 & 81.07 & 81.75 & 82.54 & 81.63 & 5.44 \\
heart-s & 82.44 & 83.09 & 81.23 & 81.73 & 74.44 & \textbf{84.07} & 78.89 & 83.09 & 79.01 & 79.51 & 3.33 \\
hepatit & 83.1 & 84.3 & 81.51 & 84.95 & 79.35 & \textbf{85.59} & 83.23 & 84.09 & 79.14 & 80.22 & 3.87 \\
hypo & 99.43 & 94.32 & \textbf{99.58} & 93.3 & 95.39 & 95.51 & 98.75 & 99.08 & 99.33 & 99.45 & 0.05 \\
iono & \textbf{92.99} & 89.84 & 90.98 & 84.43 & 83 & 83.57 & 90.79 & 92.78 & 89.84 & 90.6 & 1.71 \\
iris & 95.33 & \textbf{96.00} & 94.67 & 95.33 & 94 & 95.56 & 95.33 & 94.22 & 93.56 & 92.67 & 0.67 \\
labor & \textbf{92.98} & 87.72 & 78.36 & 89.47 & 83.63 & 92.4 & 87.72 & 85.96 & 78.36 & 82.46 & 0.00 \\
lungCan & 53.75 & 52.08 & \textbf{56.25} & 47.92 & 55.21 & 55.21 & \textbf{56.25} & 52.08 & 51.04 & 54.17 & \textbf{12.50} \\
lympho & 83.24 & \textbf{83.56} & 77.48 & 83.33 & 75.68 & 81.98 & 77.48 & 80.63 & 78.38 & 78.15 & 2.03 \\
MagicTe & 86.27 & 86.09 & 85.58 & 83.98 & 76.27 & 76.08 & 82.88 & \textbf{86.43} & 84.77 & 85.29 & 2.90 \\
nursery & \textbf{98.91} & 98.78 & 97 & 98.1 & 88.96 & 90.21 & 96.97 & 97.97 & 95.67 & 96.73 & 0.02 \\
\rowcolor{gray} ozone & 97.01 & 97.12 & 97.12 & 97.12 & 97.12 & 97.12 & 97.12 & \textbf{97.13} & 97.12 & 97.12 & 0.51 \\
pasture & 86.11 & 77.78 & 78.7 & 68.52 & \textbf{87.04} & 75 & 80.56 & 76.85 & 74.07 & 68.52 & 2.78 \\
\hline
\multicolumn{11}{c}{Continued on next page}\\
\hline
\end{tabular}
\label{table:ensembleDS}
\end{table}

\begin{table}[t]
\addtocounter{table}{-1}
\caption{{\bf (Cont.)} Comparison of the accuracy for each data set using a voting ensemble (Ens) with using an ensemble filter for the investigated learning algorithms.
The column ``Per'' refers to the percentage of instances that have a $p(\hat{y}_i|x_i)$ greater than or equal to 90\%.
The rows in gray represent those datasets where filtering with an ensemble filter increased the accuracy over the voting ensemble for all learning algorithms.}
\setlength{\tabcolsep}{0.8pt}
\begin{tabular}{l|ccccccccccc}
& Ens & MLP & C4.5 & IB5 & LWL & NB & NNge & RF & Rid & RIP & Per\\
\hline
pimaDia & \textbf{77.06} & 76.56 & 76.61 & 75.17 & 73.26 & 75.78 & 75.22 & 76.22 & 75.3 & 75.26 & 6.77 \\
\rowcolor{gray} post-op & 69.78 & \textbf{71.11} & \textbf{71.11} & \textbf{71.11} & \textbf{71.11} & \textbf{71.11} & \textbf{71.11} & \textbf{71.11} & \textbf{71.11} & \textbf{71.11} & \textbf{26.67} \\
pri-tum & 48.08 & 47.79 & 41.2 & 45.82 & 34.42 & \textbf{48.57} & 44.44 & 45.23 & 39.82 & 40.41 & \textbf{32.15} \\
segment & \textbf{98.00} & 96.05 & 96.62 & 95.04 & 78.59 & 80.69 & 96.36 & 97.37 & 95.83 & 94.82 & 0.17 \\
sick & 98.45 & 96.93 & \textbf{98.51} & 96.3 & 96.55 & 94.82 & 96.86 & 98.16 & 98.1 & 98.03 & 0.13 \\
sonar & 81.92 & 81.89 & 72.92 & \textbf{82.53} & 74.84 & 68.27 & 71.63 & 79.49 & 73.24 & 79.01 & 0.00 \\
soybean & \textbf{94.32} & 94.05 & 91.7 & 90.14 & 56.95 & 92.83 & 93.02 & 92.53 & 90.41 & 91.85 & 1.46 \\
spambas & \textbf{94.95} & 91.79 & 92.8 & 90.33 & 78.28 & 82.24 & 92.24 & 94.73 & 92.07 & 92.71 & 0.54 \\
T.A. & \textbf{57.88} & 55.19 & 51.66 & 45.25 & 50.99 & 49.89 & 52.98 & 53.42 & 43.49 & 47.9 & 8.61 \\
titanic & \textbf{78.72} & 78.66 & 78.68 & 78.59 & 77.9 & 77.77 & 78.68 & 78.68 & 78.28 & 78.68 & \textbf{16.86} \\
vote & 95.82 & 95.86 & 95.71 & 92.8 & 95.63 & 90.96 & 95.4 & \textbf{96.4} & 94.18 & 95.63 & 1.61 \\
vowel & \textbf{95.54} & 92.83 & 75.81 & 93.13 & 35.05 & 63.54 & 87.12 & 94.48 & 75.93 & 71.57 & 0.00 \\
wave & 84.21 & \textbf{85.11} & 77.92 & 79.69 & 56.93 & 79.91 & 82.44 & 81.7 & 80.11 & 79.75 & 1.34 \\
wine & 97.53 & \textbf{97.75} & 93.26 & 95.88 & 90.26 & 97.57 & 96.25 & 97.57 & 91.01 & 92.51 & 0.00 \\
yeast & \textbf{61.08} & 59.43 & 60.13 & 59.32 & 40.7 & 58.15 & 59.4 & \textbf{61.08} & 59.74 & 60.01 & \textbf{13.68} \\
zoo & 95.25 & \textbf{95.38} & 92.41 & 94.72 & 85.48 & 94.72 & 94.72 & 91.75 & 90.43 & 86.8 & 1.98 \\

\hline
Acc & 84.37 & 83.40 & 81.61 & 80.85 & 73.48 & 78.92 & 81.59 & 82.94 & 80.57 & 80.76 \\
\end{tabular}
\end{table}

\begin{table}
\caption{Comparison of the accuracy from a majority voting ensemble trained on unfiltered (Ens) and filtered data (FEns).
The value after ``FEns'' represents the percentage of learning algorithms that have to misclassify an instance for it to be filtered from the training set and ``Max'' uses the accuracy from the percentage that results in the greatest accuracy.
Training with unfiltered data is significantly better than training with filtered data.
The values in bold for the ``Ens'' represent if the majority voting ensemble is greater then the filtered majority voting ensemble.
The values in bold for the ``FEns'' columns represent if using filtered training data results in greater classification accuracy.}
\centering
\begin{tabular}{l|ccccc}
Data set & Ens & FEns 50 & FEnse 70 & FEns 90 & FEns Max \\\hline
anneal.ORIG & {\bf 98.08} & 97.57 & 96.26 & 86.15 & 97.57 \\
AP-Breast-Uterus & 97.61 & {\bf 97.69} & 97.44 & 96.15 & {\bf 97.69} \\
ar1 & 90.08 & 90.08 & {\bf 90.41} & {\bf 92.40} & {\bf 92.40} \\
arrhythmia & {\bf 71.11} & 70.40 & 69.69 & 55.58 & 70.40 \\
audiology & {\bf 78.94} & 77.52 & 72.30 & 48.58 & 77.52 \\
autos & {\bf 83.51} & 82.15 & 73.56 & 49.66 & 82.15 \\
badges2 & 100.00 & 100.00 & 100.00 & 100.00 & 100.00 \\
balance-scale & {\bf 88.45} & 86.46 & 85.28 & 71.42 & 86.46 \\
breast-cancer & 73.99 & 73.71 & {\bf 74.20} & {\bf 74.34} & {\bf 74.34} \\
breast-w & {\bf 96.88} & 96.71 & 96.62 & 93.45 & 96.71 \\
bupa & {\bf 71.30} & 70.84 & 68.35 & 60.52 & 70.84 \\
carEval & {\bf 96.70} & 95.51 & 91.81 & 70.02 & 95.51 \\
chess-KRVKP & {\bf 99.53} & 99.42 & 99.26 & 83.94 & 99.42 \\
cm1-req & 75.73 & 75.06 & {\bf 77.53} & {\bf 77.53} & {\bf 77.53} \\
colic & 85.33 & {\bf 85.54} & {\bf 85.98} & 81.52 & {\bf 85.98} \\
contact-lenses & 76.67 & {\bf 77.50} & {\bf 80.00} & 70.83 & {\bf 80.00} \\
credit-a & {\bf 86.64} & 86.26 & 86.03 & 85.51 & 86.26 \\
credit-g & {\bf 75.64} & 74.52 & 72.84 & 70.00 & 74.52 \\
dermatology & 97.43 & 97.43 & 97.27 & 91.58 & 97.43 \\
desharnais & {\bf 74.32} & 73.09 & 72.84 & 69.38 & 73.09 \\
ecoli & 87.44 & {\bf 87.74} & 86.90 & 64.88 & {\bf 87.74} \\
eucalyptus & {\bf 65.11} & 63.97 & 61.82 & 52.83 & 63.97 \\
eye-movements & {\bf 64.76} & 59.02 & 55.26 & 45.21 & 59.02 \\
glass & {\bf 74.02} & 62.52 & 61.59 & 49.53 & 62.52 \\
heart-c & {\bf 83.83} & 82.38 & 82.18 & 80.20 & 82.38 \\
heart-h & 82.93 & 82.86 & {\bf 83.40} & 82.04 & {\bf 83.40} \\
heart-statlog & {\bf 82.44} & 82.37 & 81.26 & 78.59 & 82.37 \\
hepatitis & 83.10 & {\bf 83.35} & 83.10 & 80.26 & {\bf 83.35} \\
hypothyroid & {\bf 99.43} & 99.37 & 98.17 & 94.04 & 99.37 \\
ionosphere & {\bf 92.99} & 92.82 & 91.34 & 84.67 & 92.82 \\
iris & {\bf 95.33} & 94.53 & 94.13 & 94.13 & 94.53 \\
labor & {\bf 92.98} & 91.58 & 88.07 & 82.11 & 91.58 \\
lungCancer & {\bf 53.75} & 51.25 & 53.13 & 38.75 & 53.13 \\
lymphography & {\bf 83.24} & 81.35 & 80.95 & 76.35 & 81.35 \\
MagicTelescope & {\bf 86.27} & 85.49 & 84.73 & 74.91 & 85.49 \\
nursery & {\bf 98.91} & 98.55 & 97.24 & 90.43 & 98.55 \\
\hline
\multicolumn{6}{c}{Continued on next page}\\
\hline
\end{tabular}
\label{table:ensVsFilteredEnsDSs}
\end{table}

\begin{table}
\addtocounter{table}{-1}
\caption{{\bf Cont.} Comparison of the accuracy from a majority voting ensemble trained on unfiltered (Ens) and filtered data (FEns).
The value after ``FEns'' represents the percentage of learning algorithms that have to misclassify an instance for it to be filtered from the training set and ``Max'' uses the accuracy from the percentage that results in the greatest accuracy.
Training with unfiltered data is significantly better than training with filtered data.
The values in bold for the ``Ens'' represent if the majority voting ensemble is greater then the filtered majority voting ensemble.
The values in bold for the ``FEns'' columns represent if using filtered training data results in greater classification accuracy.}
\centering
\begin{tabular}{l|ccccc}
Data set & Ens & FEns 50 & FEnse 70 & FEns 90 & FEns Max \\\hline
ozone & 97.01 & {\bf 97.07} & {\bf 97.09} & {\bf 97.12} & {\bf 97.12} \\
pasture & {\bf 86.11} & 81.11 & 78.89 & 66.67 & 81.11 \\
pimaDiabetes & {\bf 77.06} & 76.46 & 75.81 & 73.91 & 76.46 \\
post-opPatient & 69.78 & {\bf 70.22} & {\bf 71.11} & {\bf 71.11} & {\bf 71.11} \\
primary-tumor & {\bf 48.08} & 45.19 & 43.01 & 38.47 & 45.19 \\
segment & {\bf 98.00} & 97.62 & 96.54 & 85.65 & 97.62 \\
sick & {\bf 98.45} & 98.32 & 98.17 & 97.16 & 98.32 \\
sonar & {\bf 81.92} & 81.44 & 80.10 & 73.75 & 81.44 \\
soybean & {\bf 94.32} & 93.85 & 93.12 & 67.88 & 93.85 \\
spambase & {\bf 94.95} & 94.78 & 94.17 & 84.43 & 94.78 \\
teachingAssistant & {\bf 57.88} & 54.44 & 47.15 & 39.60 & 54.44 \\
titanic & {\bf 78.72} & 78.65 & 78.00 & 77.60 & 78.65 \\
vote & {\bf 95.82} & 95.72 & 95.68 & 95.40 & 95.72 \\
vowel & {\bf 95.54} & 94.75 & 86.44 & 40.14 & 94.75 \\
waveform-5000 & 84.21 & {\bf 84.26} & 81.77 & 63.42 & {\bf 84.26} \\
wine & 97.53 & {\bf 97.64} & 96.97 & 96.18 & {\bf 97.64} \\
yeast & {\bf 61.08} & 61.01 & 60.55 & 40.50 & 61.01 \\
zoo & {\bf 95.25} & 94.65 & 93.66 & 87.13 & 94.65 \\\hline
Ave & 84.37 & 83.40 & 82.21 & 73.96 & 83.62 \\
\end{tabular}

\end{table}

\end{document}